\documentclass[lettersize,9pt,journal]{IEEEtran}
\usepackage{amsmath,amsfonts}
\usepackage{algorithm}
\usepackage{array}
\usepackage[caption=false,font=normalsize,labelfont=sf,textfont=sf]{subfig}
\usepackage{textcomp}
\usepackage{stfloats}
\usepackage{url}
\usepackage{verbatim}
\usepackage{cite}
\usepackage{graphicx} 
\usepackage{epsfig} 
\usepackage{tensor}
\usepackage{siunitx}

\usepackage{color}
\usepackage{times}
\usepackage{balance}
\usepackage{fancyhdr}
\usepackage{comment}
\usepackage{changepage}
\usepackage{bm}
\usepackage{calligra}
\usepackage{tabularx}
\usepackage{mathtools}
\usepackage{arydshln}
\usepackage{latexsym} 
\usepackage{float}
\usepackage{tabstackengine}
\usepackage{threeparttable}
\usepackage{subcaption}
\usepackage{xcolor}
\usepackage{booktabs}
\usepackage{algpseudocode}
\usepackage{multirow}
\usepackage{colortbl} 
\usepackage{amssymb}
\definecolor{lightgray}{rgb}{0.9, 0.9, 0.9} 
\usepackage{pifont} 

\usepackage{hyperref}
\hypersetup{
    colorlinks=true,
    linkcolor=subsectioncolor,
    filecolor=magenta,      
    urlcolor=cyan,
}

\definecolor{subsectioncolor}{RGB}{255,0,0} 

\newcommand{\rt}{\textcolor{red}}

\begin{document}

\title{MS-Mapping: An Uncertainty-Aware Large-Scale Multi-Session LiDAR Mapping System
}

\author{Xiangcheng Hu,~\IEEEmembership{Student Member,~IEEE,}

\author{
Xiangcheng Hu$^{1}$,
Jin Wu$^{1}$,
Jianhao Jiao$^{2}\dagger$, 
Binqian Jiang$^{1}$, 
Wei Zhang$^{1}$,
Wenshuo Wang$^{3}$
and Ping Tan$^{1}\dagger$

\thanks{$^{1}$X. Hu, J. Wu, B. Jiang, W. Zhang and P. Tan are with the Department of Electronic and Computer Engineering, Hong Kong University of Science and Technology, Hong Kong, China (E-mail: pingtan@ust.hk)
}
\thanks{$^{2}$J. Jiao is with the Department of Computer Science, University College London, Gower Street, WC1E 6BT, London, UK. (E-mail: ucacjji@ucl.ac.uk)
}
\thanks{$^3$W. Wang is with Department of Mechanical Engineering at Beijing Institute of Technology (BIT). (E-mail: wwsbit@gmail.com)}
\thanks{$^{\dagger}$Corresponding Author}
}

}%


\markboth{Submitted to IEEE Transactions on Robotics},

\maketitle

\begin{abstract}

Large-scale multi-session LiDAR mapping is essential for a wide range of applications, including surveying, autonomous driving, crowdsourced mapping, and multi-agent navigation. However, existing approaches often struggle with data redundancy, robustness, and accuracy in complex environments. To address these challenges, we present MS-Mapping, an novel multi-session LiDAR mapping system that employs an incremental mapping scheme for robust and accurate map assembly in large-scale environments. Our approach introduces three key innovations:
1) A distribution-aware keyframe selection method that captures the subtle contributions of each point cloud frame to the map by analyzing the similarity of map distributions. This method effectively reduces data redundancy and pose graph size, while enhancing graph optimization speed;
2) An uncertainty model that automatically performs least-squares adjustments according to the covariance matrix during graph optimization, improving mapping precision, robustness, and flexibility without the need for scene-specific parameter tuning.
This uncertainty model enables our system to monitor pose uncertainty and avoid ill-posed optimizations, thereby increasing adaptability to diverse and challenging environments.
3) To ensure fair evaluation, we redesign baseline comparisons and the evaluation benchmark. Direct assessment of map accuracy demonstrates the superiority of the proposed MS-Mapping algorithm compared to state-of-the-art methods.
In addition to employing public datasets such as Urban-Nav, FusionPortable, and Newer College, we conducted extensive experiments on such a large \SI{855}{m}$\times$\SI{636}{m} ground truth map, collecting over \SI{20}{km} of indoor and outdoor data across more than ten sequences. These comprehensive experiments highlight the robustness and accuracy of our approach.
To facilitate further research and development in the community, we  would make our code \footnote{https://github.com/JokerJohn/MS-Mapping} and datasets\footnote{https://github.com/JokerJohn/MS-Dataset}  publicly available.

\end{abstract}



%


\section{Introduction}

\subsection{Motivation}

\begin{figure}
    \centering
    \includegraphics[width=0.5\textwidth]{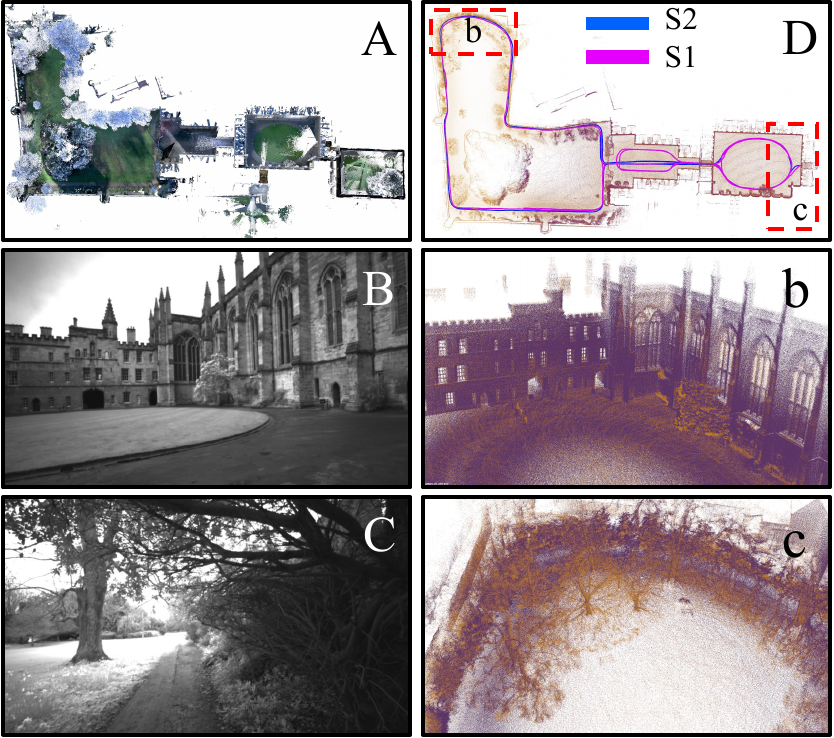}
    \caption{
    \footnotesize
  Visualization of incremental mapping results using the MS-mapping algorithm with the \texttt{short-exp11} (\texttt{S2}) and the \texttt{parkland0} sequence (\texttt{S1}) (Table~\ref{tab:algorithm-comparison-nc}). (A) Ground truth map. (D) Merged map and trajectory. (B)-(C) Real-world images of regions b and c, where baseline algorithms show significant errors. (b)-(c) Point cloud map of regions a and b in (D).
  }
    \label{fig:in_map_vi}
    \vspace{-1.5em}
\end{figure}

\IEEEPARstart{T}he rapid advancement of LiDAR Simultaneous Localization and Mapping (SLAM) has revolutionized 3D perception and navigation capabilities in robotics and autonomous systems. While single-session LiDAR SLAM\cite{shan2021liosam,xu2022fast,ye2019tightly} has achieved remarkable success in creating detailed 3D representations of environments, its limitations become evident as the scale and complexity of mapping tasks increase. Multi-session LiDAR mapping\cite{yin2023auto} addresses these limitations and unlocks new possibilities in various applications.
Fig.~\ref{fig:in_map_vi} illustrates a example of multi-session mapping on the Newer College dataset\cite{zhang2021multicamera}. Using new session data, we enhance and refine the existing map built from old session data.
By integrating data from multiple sessions captured at different times and locations, multi-session LiDAR mapping enables seamless spatial expansion and refinement of 3D maps. This approach allows for large-scale mapping without requiring a complete re-survey when data deviates from the expected collection route. Additionally, it reduces the extensive workload associated with large-scale mapping projects by permitting targeted and efficient supplementary data collection.
Multi-session LiDAR mapping is becoming increasingly crucial in several key application domains. In surveying and urban planning, it facilitates the creation of expansive, high-fidelity digital twins of cities and infrastructure, essential for informed decision-making and efficient resource management. For autonomous driving, multi-session mapping provides richer and more accurate road models, enhancing navigation safety and efficiency in complex urban environments. In multi-agent exploration, such as search and rescue operations, it enables the creation of a shared, consistent environmental model, crucial for effective coordination among multiple agents.
Furthermore, multi-session mapping serves as a crucial foundation for lifelong SLAM \cite{kim2022lt} and semantic mapping. 
By providing an accurate and up-to-date base map, multi-session mapping enables these downstream tasks to be performed more flexibly and efficiently.
Given these considerations, developing efficient and accurate multi-session LiDAR mapping algorithms is a necessary evolution in the field of robotics navigation.

\subsection{Challenges}


Despite the advantages of multi-session mapping algorithms in large-scale environments, several significant issues persist:
\subsubsection{Data Redundancy and Computational Efficiency}
Large-scale LiDAR mapping generates massive point cloud data, which leads to significant redundancy, high memory consumption, and computationally inefficiency. These issues often result in severe performance degradation and hinder long-term scalability, particularly on resource-constrained platforms.
Therefore, developing efficient keyframe selection algorithm is crucial for reducing data redundancy while enhancing optimization speed.

\begin{figure}
    \centering
    \includegraphics[width=0.4\textwidth]{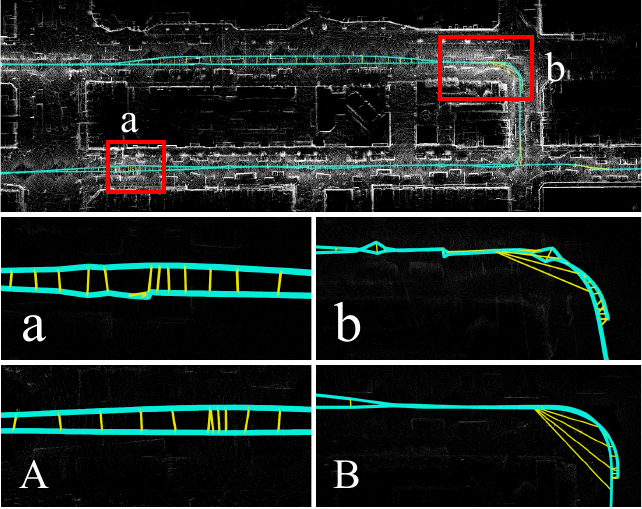}
    \caption{
    \footnotesize 
Frequent appearance of ill-posed graph optimization in large-scale SLAM. 
The top subfigure shows the point cloud map of the \texttt{Mongkok} sequence\cite{hsu2023hong} constructed using the FPGO. Regions a and b illustrate how pose trajectories are distorted during graph optimization with incorrect noise settings in dense loop closure scenarios, which lead to ill-posed optimization and mapping failure. Properly modeled uncertainty, as shown in subfigures A and B, allows the system to effectively optimize these areas.
    }
    \label{fig:ill_pose}
        \vspace{-1.5em}
\end{figure}

\subsubsection{Map Accuracy and Robustness}\label{sub_sec:chall_acc}

Achieving high accuracy and consistency in the merged map, particularly in overlapping regions, remains a critical challenge in multi-session mapping\cite{yin2023auto}. 
Pose graph optimization-based approaches (PGO), although widely adopted, can result in poor global consistency and ghost artifacts due to insufficient or low-quality loop closures\cite{chen2022fast}. 
Fixed-covariance pose graph optimization (FPGO), despite its widespread use\cite{shan2021liosam}, assumes fixed-covariance for loop closure and odometry edges, which contradicts standard optimization theory. 
In long-term SLAM, this assumption fails to accurately reflect the quality of relative poses and loop closure constraints, leading to significant accuracy degradation and map ghosting.
Furthermore,  the assumptions of FPGO can lead to ill-posed optimizations in large-scale mapping (see Fig.~\ref{fig:ill_pose}), a problem known as pose graph relaxation\cite{barfoot2024state} and the Gauge Freedom problem\cite{zhang2018comparison}. 
To achieve relatively good optimization results, FPGO requires scene-specific parameter tuning of the covariance matrix for various edges. However, this tuning mechanism cannot fundamentally solve the problem. 
Moreover, this approach hinders meaningful localization uncertainty estimation, posing challenges for safe navigation.


\subsubsection{Unfair Experimental Benchmarks}

Existing multi-session mapping algorithms primarily focus on localization experiments\cite{yuan2023lta}, while neglect map evaluation and the usage of datasets with ground truth maps. Morever, these studies either lack system baselines or solely compare against disparate LiDAR SLAM algorithms, making it difficult to pinpoint the source of improvements between odometry and backend optimization. This limitation hinders the quantification of overall advancements in multi-session mapping algorithms.


\subsection{Contribution}
The proposed MS-Mapping system emphasizes three key aspects:
\begin{itemize}
    \item We introduce a distribution-aware keyframe selection method that captures the subtle changes each point cloud frame contributes to the map based on the similarity of map distributions. This is achieved by quantifying the differences between two distributions based on Wasserstein distance, which effectively reducing data redundancy and enhance graph optimization speed.
    \item We present an uncertainty model that automatically performs least-squares adjustments during graph optimization, improving mapping precision and robustness without the need for scene-specific parameter tuning. This model also monitors pose uncertainty and avoids ill-posed optimizations.
    \item We develop a comprehensive evaluation benchmark for multi-session LiDAR mapping tasks, including map assessments, implemented baseline algorithms, and open-sourced datasets with a large-scale indoor and outdoor ground truth map to facilitate further research and development in the community.
\end{itemize}

\subsection{Outline}


The paper is organized as follows. Section~\ref{sec:related_work} reviews related work in multi-session LiDAR mapping. Section~\ref{sec:system_overview} provides an overview of the MS-Mapping system architecture. Section~\ref{sec:ws_keyframe} explains the keyframe selection method, while Section~\ref{sec:uncertainty_slam} details the uncertainty model in SLAM. Section~\ref{sec:dataset} describes the datasets and experimental design. Sections~\ref{sec:module_exp} and \ref{sec:sec_sys_exp} present the experimental results for MS-Mapping. Finally, Section~\ref{sec:conclusion} concludes the paper.

\section{Related Work}\label{sec:related_work}


This section reviews pertinent literature on multi-session LiDAR mapping, focusing on backend optimization and system-level aspects. Although place recognition \cite{taubner2020lcd,kim2021scan,jiang2023contour} is a core challenge in multi-session SLAM, it is not reviewed in this paper.

\subsection{LiDAR Keyframe Selection}
The massive volume of LiDAR point cloud data presents significant challenges in terms of data redundancy \cite{schmuck2019redundancy}, memory consumption, and computational efficiency. 
While pose graph pruning \cite{wang2015pose,kurz2021geometry} or sparsification techniques \cite{nam2023spectral} offer offline solutions, their applicability in real-time scenarios is limited. This limitation stems from their requirement to identify optimal combinations from complete pose graph sets to characterize entire scenes while minimizing accuracy loss and information degradation.
Keyframe selection emerges as an alternative approach to reduce data redundancy and pose graph size, indirectly enhancing graph optimization speed. These methods generally fall into two categories: scene-aware and motion state-based approaches.
Scene-aware \cite{alonso2019keyfr} or information theory-based techniques \cite{das2015entropy} typically compute covariance matrices from a large number of environmental feature points to define information gain or residuals, thereby selecting optimal frame combinations as keyframe sets. While these methods directly consider scene changes, they often lack real-time performance due to computational burdens.
Conversely, methods based on motion states (distance or angle thresholds) \cite{ye2019tightly} require minimal additional computation. However, this brute-force selection approach often leads to information loss and incomplete mapping.
To address these challenges, we extend the scene-aware approach, aiming for a more comprehensive representation of map information. We propose modeling the LiDAR keyframe selection problem as a similarity measure between Gaussian mixture distributions (GMM). By observing the map distribution before and after adding a point cloud frame, we employ a Wasserstein distance-based method to capture both global and local differences. To enhance efficiency, we divide the map into voxels and utilize an incremental voxel update method for Gaussian parameter updates in the GMM map.

\subsection{Uncertainty LiDAR SLAM}\label{sub_sec:related_uncer}

While uncertainty modeling has been extensively studied in visual SLAM\cite{qin2018vins} and state estimation\cite{barfoot2024state}, its application in LiDAR SLAM remains limited. This disparity stems from the unique characteristics of LiDAR sensor : high measurement accuracy, direct depth measurements, and numerous feature points. These factors often lead to indirect map optimization through pose optimization, creating discrepancies with standard graph optimization theory regarding pose and map covariance. Nevertheless, real-time pose uncertainty estimation is also crucial for safe robot decision and navigation in some exploration tasks\cite{yang2024efficient,zhou2023racer }.

Theoretical works have laid foundations for uncertainty modeling in Pose-SLAM. Some assume pose independence\cite{barfoot2014associating}, while others further employ high-order approximations using the BCH formula\cite{mangelson2020characterizing} or quaternion parameterization for global optimal pose covariance estimation\cite{wu2022quadratic}. However, these methods often struggle with improved performance in real-world LiDAR SLAM applications, with most validations limited to $SE(2)$ optimization or simulation experiments.
Most algorithms employ fixed-covariance graph optimization, assuming constant covariance for loop closure and odometry edges. This assumption, while successful in small-scale scenarios, can lead to significant issues in large-scale mapping. 
As mapping scale expands and loop closure constraints increase, pose graph relaxation problem\cite{barfoot2024state} may arise, necessitating scene-specific parameter tuning or mapping failure.
The covariance of the initial prior factor, introduced during pose graph initialization, can significantly impact optimization stability, a phenomenon known as the Gauge Freedom problem\cite{zhang2018comparison}.
Moreover, fixed-covariance optimization precludes meaningful localization uncertainty estimation, challenging safe navigation.
Notable LiDAR SLAM system\cite{shan2018lego, shan2021liosam} employ fixed-covariance noise models for both odometry and loop closure edges, necessitating scene-specific parameter tuning in large-scale scenarios and risking ill-posed optimizations. 
M-LOAM\cite{jiao2021robust} proposed an optimization-based multi-LiDAR online calibration and odometry algorithm considering zero-mean Gaussian noise for LiDAR feature points, but lacked loop closure capabilities. 
Xu et al.\cite{xu2022fast} developed FAST-LIO2, which implements continuous two-frame pose covariance estimation within an Iterative Error-state Kalman Filter (IEKF). In a similar vein, Yuan et al.~\cite{yuan2022efficient} modeled noise based on point incidence angles and depths, although their method is limited to mechanical LiDARs and does not consider the influence of Inertial Measurement Units (IMUs).

One of our recent work\cite{chen2022fast} explored Bayesian ICP\cite{maken2020estimating} for loop closure covariance estimation but did not consider the relative scale of odometry covariance. Further, we applied uncertainty models in offline localization systems\cite{hu2024paloc}, though limited to small-scale scenarios without comprehensive experimental analysis.
In this work, we introduce a uncertainty model that reshapes the entire LiDAR SLAM process. Our approach derives covariance for both odometry and loop closure edges, ensuring consistent scale across all pose graph factors. Our new method significantly improves accuracy and robustness in large-scale scenarios while mitigating ill-posed optimization issues.

\subsection{Multi-session LiDAR Mapping}\label{sub_sec:related_multi}

Multi-session SLAM, primarily focused on short-term mapping with static environmental assumptions, has been extensively studied in visual SLAM\cite{campos2021orb, labbe2019rtab,schneider2018maplab,tian2022kimera,dube2020segmap,cramariuc2021semsegmap,lajoie2020door,karrer2018cvi}. However, our focus is on multi-session LiDAR mapping, where achieving high accuracy and consistency in merged maps, particularly in overlapping regions, remains a critical challenge\cite{yuan2023lta}.

LiDAR bundle adjustment\cite{liu2023efficient,huang2021bundle,liu2023hba} methods offer direct map optimization but face limitations in efficiency especially in large-scale applications. These methods require extreme pose graph sparsification and extensive offline processing, resulting in incomplete maps and reduced applicability to large-scale scenes. Moreover, their planar constraints often struggle in outdoor environments with significant plane noise.
Consequently, pose graph optimization-based approaches remain preferable for large-scale multi-session mapping. 
Giseop et al.\cite{kim2022lt}, the first complete lifelong LiDAR mapping framework, introduced adaptive covariance for inlier loop closures but suffered from accuracy degradation due to fixed odometry covariance and inconsistent edge weighting, overlooking the relative nature of covariance in pose graph (Section~\ref{sub_sec:related_uncer}). Automerge\cite{yin2023auto} demonstrated large-scale map merging capabilities but lacked efficiency considerations and comprehensive baseline comparisons. 
Cramariuc et al.\cite{cramariuc2022maplab} extended their previous work, \texttt{Maplab}, to support multi-session, multi-robot LiDAR mapping in \texttt{Maplab 2.0}, although their core contributions were primarily engineering-focused. 
Zou et al.\cite{yuan2023lta} proposed LTA-OM, which relied on localization in existing maps, limiting its accuracy to prior map precision.
\cite{sta2024frame} developed FRAME, a system that utilizes place recognition and learned descriptors to efficiently detect overlap, followed by the GICP algorithm for map matching.
While previous multi-robot SLAM approaches~\cite{chang2022lamp,mangelson2018pairwise,pattabiraman2015fast} have focused on developing robust back-ends with outlier-resilient pose graph optimization, these methods prioritize high real-time localization. Place recognition algorithms, such as BTC~\cite{yuan2024btc} and Ring++~\cite{xu2023ring}, have been applied to multi-session mapping, but they primarily focus on evaluating place recognition algorithms and assess system accuracy solely through localization experiments.


A significant limitation in existing algorithms is that they lack direct map evaluation, with the exception of \cite{kim2022lt}, which employed a simple Chamfer distance for map evaluation. While localization accuracy can indirectly reflect map quality, direct map evaluation provides a more straightforward way.
Furthermore, the lack of fair baselines and comparisons against diverse LiDAR SLAM algorithms obscures the source of improvements in these systems.
Recent SLAM datasets with dense ground truth maps\cite{jiao2022fusionportable, wei2024fusionportablev2,nguyen2022ntu,ramezani2020newer,zhang2021multicamera}, combined with our map evaluation framework\cite{hu2024paloc}, have enabled us to develop a comprehensive benchmark and implement baseline algorithms. This approach facilitates thorough and fair map evaluation, providing clear insights into accuracy improvements and enabling meaningful cross-method comparisons.

\begin{figure*}
    \centering
    \includegraphics[width=0.8\textwidth]{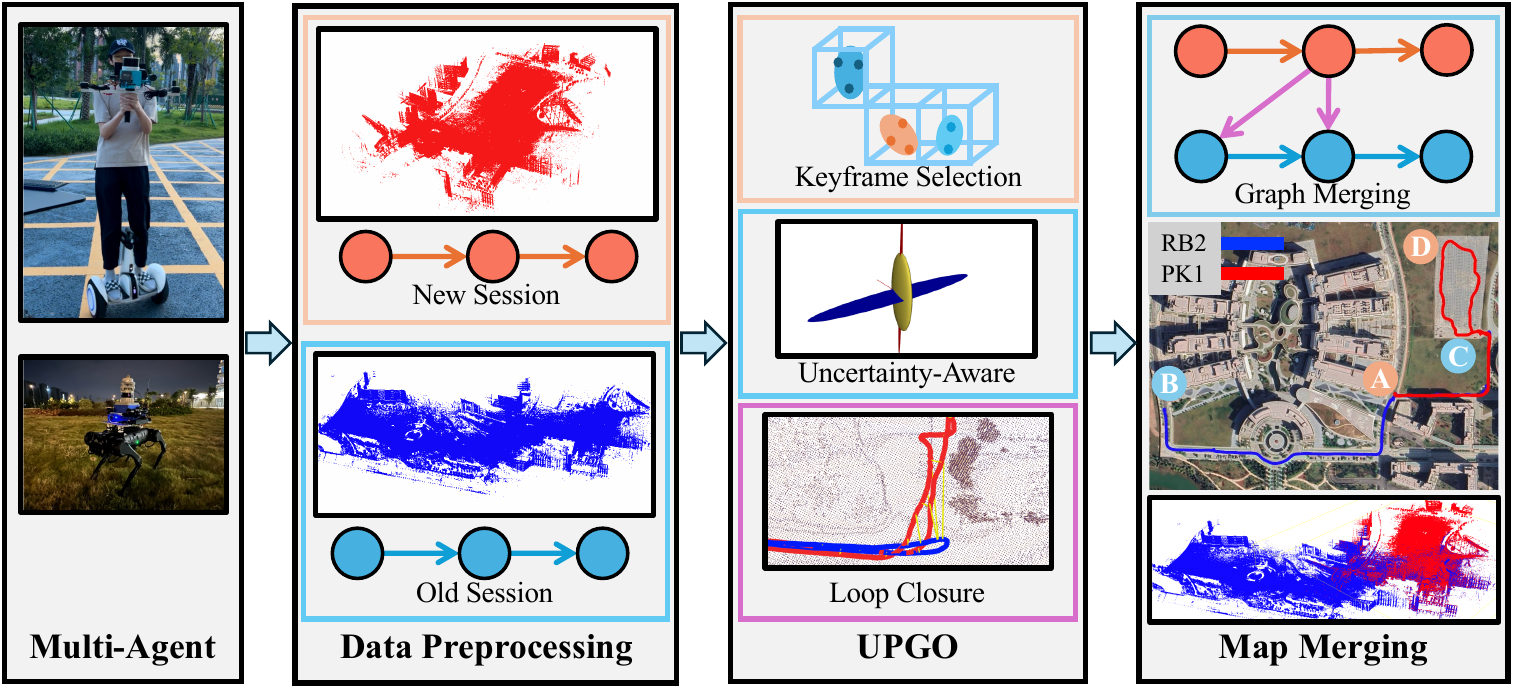}
    \caption{
    \footnotesize
    \textbf{Overview of system architecture}. 
    Our MS-Mapping system integrates data from distinct agents. The process begins with constructing a pose graph and map frames for the old session (\texttt{RB2}, path: A$\rightarrow$C$\rightarrow$A$\rightarrow$B). Subsequently, we incrementally update the map with data from a new session (\texttt{PK1}, path: A$\rightarrow$C$\rightarrow$D$\rightarrow$A). Key frames are selected based on their contribution to the map using our distribution-aware method. The system performs UPGO to produce a merged pose graph and point cloud map. The right panel illustrates the projection of the final trajectory onto satellite maps.
    }
    \label{fig:pipeline}
        \vspace{-1.5em}
\end{figure*}

\section{Problem Formulation}

\subsection{Notations}\label{subsec:notations}
Let $\mathcal{W}$ denote the common world frame. The transformation from the body frame $\mathcal{B}$ to the world frame at time instant $i$ is represented as:
$
\small
\mathbf{T}_{b}^w = \begin{bmatrix}
\mathbf{R}_{b}^w & \mathbf{t}_{b}^w \\
\mathbf{0} & 1
\end{bmatrix} \in SE(3),$
where $\mathbf{R}_{b_i}^w \in SO(3)$ is the rotation matrix and $\mathbf{t}_{b_i}^w \in \mathbb{R}^3$ is the translation vector. We adhere to the right-hand convention for manifold derivatives, to be consistent with the optimization library GTSAM\cite{dellaert2012factor}. This implies that a small perturbation on the manifold is expressed in the body frame, which can be converted to the world frame using the concept of adjoint, denoted as $\text{Ad}_{(\cdot)}$.
Given two data sequences $\mathcal{S}_1$ and $\mathcal{S}_2$,
we can first employ LiDAR-based odometry algorithms to obtain corresponding pose trajectories and de-skewed point cloud frame sets. 
We define their pose trajectories as $\mathcal{X}_1 = \{\mathbf{T}_1^{w_1}, \mathbf{T}_2^{w_1}, \dots, \mathbf{T}_n^{w_1}\}$ and $\mathcal{X}_2 = \{\mathbf{T}_1^{w_2}, \mathbf{T}_2^{w_2}, \dots, \mathbf{T}_m^{w_2}\}$. The corresponding point cloud set are denoted as $\mathcal{P}_1 = \{{}^1\mathbf{P}_1, {}^1\mathbf{P}_2, \dots, {}^1\mathbf{P}_n\}$ and $\mathcal{P}_2 = \{{}^2\mathbf{P}_1, {}^2\mathbf{P}_2, \dots, {}^2\mathbf{P}_m\}$, where the left superscript indicates the data sequence and the subscript represents the frame index. The merged pose graph is represented as $\mathcal{G} = (\mathcal{V}, \mathcal{E})$, where $\mathcal{V} = \mathcal{X}_1 \cup \mathcal{X}_2$ is the set of pose nodes and $\mathcal{E}$ is the set of edges, including both odometry edges and loop closure edges.

\subsection{Pose-SLAM With Uncertainty Modeling }\label{subsec:prior_map_assisted_pgo}
In multi-session mapping, our objective is to expand the pose graph and construct a globally consistent map by optimizing the poses of $\mathcal{S}_2$ based on the existing pose graph of $\mathcal{S}_1$. This is achieved by incorporating LiDAR odometry constraints and loop closure between the two data sequences.
A fundamental assumption in multi-session mapping is the existence of overlapping regions between the two sequences, which can be detected using place recognition algorithms. 
Therefore we assume that an initial pose estimate between the sequences is known, allowing us to reformulate the multi-session mapping task as a Pose-SLAM problem with loop constraints in the overlapping regions.
Given an initial guess $\mathbf{T}_{init}$ between $\mathcal{S}_1$ and $\mathcal{S}_2$, we aim to estimate the optimal merged pose trajectory $\mathcal{V}$ that minimizes the overall error in the merged pose graph $\mathcal{G}$:
\begin{equation}\label{eq:uncertain_slam}
\arg\min_{\mathcal{V}} \sum_{i \in \mathcal{E}_{od}^{\mathcal{G}}} \|\mathbf{e}_{od_i}^{\mathcal{G}}\|_{\mathbf{\Omega}_{od}}^2 + \sum_{j \in \mathcal{E}_{lc}^{\mathcal{G}}} \|\mathbf{e}_{lc_j}^{\mathcal{G}}\|_{\mathbf{\Omega}_{lc}}^2 + \sum_{k \in \mathcal{E}_{pr}^{\mathcal{G}}} \|\mathbf{e}_{pr_k}^{\mathcal{G}}\|_{\mathbf{\Omega}_{pr}}^2,
\end{equation}
where $\mathcal{E}_{od}^{\mathcal{G}}$, $\mathcal{E}_{lc}^{\mathcal{G}}$, and $\mathcal{E}_{pr}^{\mathcal{G}}$ represent the sets of odometry edges, loop closure edges, and prior factors in the merged pose graph $\mathcal{G}$. The corresponding edge errors are denoted as $\mathbf{e}_{od_i}^{\mathcal{G}}$, $\mathbf{e}_{lc_j}^{\mathcal{G}}$, and $\mathbf{e}_{pr_k}^{\mathcal{G}}$. The information matrix for each type of error is denoted by $\mathbf{\Omega}_{od}$, $\mathbf{\Omega}_{lc}$, and $\mathbf{\Omega}_{pr}$.
The weighted residuals are defined as:
\begin{equation}
\|\mathbf{e}_{i}(\mathbf{x})\|_{\mathbf{\Omega}_{i}}^2 = \mathbf{e}_{i}(\mathbf{x})^\top \mathbf{\Omega}_{i} \mathbf{e}_{i}(\mathbf{x}).
\end{equation}
Typically, there is only one prior residual, which is the pose residual at the initial time step used to fix the world coordinate system. Odometry residuals are the most numerous, while loop closure relative pose residuals are relatively fewer.

In practice, a common approach assumes constant covariance for all residuals, implying equal importance for all loop closure and odometry constraints. This is represented by diagonal Gaussian noise models for each type of residual. Specifically, the noise matrices $\mathbf{\Lambda}_{od}$, $\mathbf{\Lambda}_{lc}$, and $\mathbf{\Lambda}_{pr}$ are diagonal matrices of the form:
\[
\mathbf{\Lambda} = \mathrm{diag}(\lambda_r, \lambda_r, \lambda_r,\lambda_t, \lambda_t, \lambda_t),
\]
where $\lambda_r$ and $\lambda_t$ represent the noise for rotation and translation. The specific noise matrices $\mathbf{\Lambda}_{od}$, $\mathbf{\Lambda}_{lc}$, and $\mathbf{\Lambda}_{pr}$ represent odometry, loop closure, and prior factors. The cost function then simplifies to:
\begin{equation}\label{eq:simplified_slam}
\arg\min_{\mathcal{V}} \sum_{i \in \mathcal{E}_{od}^{\mathcal{G}}} \|\mathbf{e}_{od_i}^{\mathcal{G}}\|_{\mathbf{\Lambda}_{od}}^2 + \sum_{j \in \mathcal{E}_{lc}^{\mathcal{G}}} \|\mathbf{e}_{lc_j}^{\mathcal{G}}\|_{\mathbf{\Lambda}_{lc}}^2 + \sum_{k \in \mathcal{E}_{pr}^{\mathcal{G}}} \|\mathbf{e}_{pr_k}^{\mathcal{G}}\|_{\mathbf{\Lambda}_{pr}}^2.
\end{equation}
However, this assumption is unrealistic in large-scale scenarios:
\begin{enumerate}
    \item The precision of odometry constraints varies with different environmental conditions.
    \item Loop closure pose errors from iterative closest point (ICP) algorithms are typically higher than odometry errors due to less accurate initial values.
    \item In large-scale environments, loop closure measurement noise varies with the scene and initial pose, making the constant noise assumption unrealistic.
\end{enumerate}
By properly modeling the noise of the above constraints, we can achieve more accurate results and avoid potential ill-posed optimizations. This approach, termed uncertainty SLAM (UPGO), which will be detailed in Section~\ref{sec:uncertainty_slam}. This is crucial in LiDAR SLAM, where measurement noise can vary significantly based on sensor characteristics, environmental conditions, and the observed scenes.
The resulting weighted nonlinear optimization problem can further be solved using iterative methods such as Gauss-Newton or Levenberg-Marquardt algorithms. This uncertainty-aware formulation allows the optimization process to adjust the influence of different constraints based on their reliability, leading to more accurate and robust mapping results. Finally, pose covariance can be derived from the global information matrix after optimization \cite{sola2018micro,hu2024paloc}.

\section{System Overview}\label{sec:system_overview}

In this section, we present the map merging process of MS-Mapping, as illustrated in Fig.~\ref{fig:pipeline}. The MS-Mapping integrates multi-sensor data collected from multiple agents for accurate map merging. The process consists of three main steps: data preprocessing, UPGO, and map merging.  
As UPGO is the primary focus of this paper, we will provide a detailed explanation of the distribution-aware keyframe selection module in Section~\ref{sec:ws_keyframe} and discuss the uncertainty model for LiDAR SLAM in Section~\ref{sec:uncertainty_slam}. The data preprocessing and map merging modules will not be elaborated upon.

During the data preprocessing phase, our proposed single-session uncertainty SLAM system constructs a pose graph and point cloud map from the old session data. Given the initial pose of the starting point in the new session, we also estimate the online odometry of the new session.
In the UPGO phase, we first use a distribution-aware keyframe selection module to calculate the contribution of each point cloud frame to the map, selecting keyframes that significantly impact the map. 
This reduces the data redundancy and pose graph size in the backend. Subsequently, keyframes from the new session are used for loop closure detection and constructing relative pose constraints. The online odometry of the new session also provides adjacent pose constraints to the backend. Our uncertainty-aware approach accurately measures the noise of loop closure constraints and odometry relative pose constraints, maintaining the prior noise of starting pose on the identical scale. This enables the robust and accurate expansion of the pose graph.
In the map merging phase, we optimize the expanded pose graph to obtain poses and their corresponding keyframe point clouds, achieving map merging.
Fig.~\ref{fig:pipeline} illustrates this process of incremental mapping using the \texttt{PK1} sequence (red trajectory) on the \texttt{RB2} sequence (blue trajectory).

\begin{figure}
    \centering
    \includegraphics[width=0.4\textwidth]{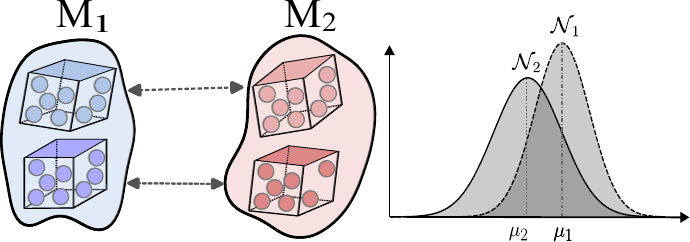}
\caption{
    \footnotesize
Evaluation of discrepancies between point cloud maps $M_1$ and $M_2$ using the Wasserstein distance in (\ref{eq:wasserstein_distance}). This metric captures global and local distribution differences. After voxelization, the point count of each voxel denotes its mass. The Wasserstein distances between voxel pairs are averaged to represent the overall difference between GMM maps.
}
\label{fig:waserstain}
\vspace{-1.5em}
\end{figure}

\section{Distribution-aware Keyframe Selection}\label{sec:ws_keyframe}
In this section, we provide a detailed explanation of the first step in UPGO: the distribution-aware keyframe selection method.

\subsection{Keyframe Selection Formulation}\label{sec:sub_keyframe}

We formulate the keyframe selection problem as a map information gain maximization task. Intuitively, if a point cloud frame $P$, when added to the map $M$, can bring significant map changes, it will be considered as a keyframe. Mathematically, let $M_1$ and $M_2$ denote the map distributions before and after the update of the new frame. 
The keyframe selection problem can be transformed into measuring the dissimilarity between two distributions, as shown in $\text{KF} = \textbf{Dis}(M_1, M_2)$,
where $\textbf{Dis}(\cdot)$ represents the distance function between two map distributions $M_1$ and $M_2$.
The map distribution can be represented as a Gaussian Mixture Model:
\begin{equation*}
M = \sum_{i=1}^{K} w_i \cdot \mathcal{N}(\mathbf{\mu_i}, \mathbf{\Sigma_i}),
\end{equation*}
\noindent where $K$ is the number of components, $w_i$ is the weight of the $i$-th component, and $\mathcal{N}(\mathbf{\mu_i}, \mathbf{\Sigma_i})$ is the Gaussian distribution with mean $\mathbf{\mu_i}$ and covariance $\mathbf{\Sigma_i}$.
When a new point cloud frame $P$ is available, it is first transformed into the map coordinate system using the estimated pose $\mathbf{T} \in SE(3)$, $P_M = \mathbf{T} \cdot P$.
To guarantee the reliability of the keyframe selection process, we make two assumptions:
\begin{enumerate}
\item The pose $\mathbf{T}$ and the map $M$ are accurately estimated.
\item The map $M$ completely covers the range of the transformed frame $P_M$.
\end{enumerate}
The first assumption ensures a consistent scale between $M_1$ and $M_2$, while the second one avoids the interference of newly explored areas on the map information gain.

\subsection{Wasserstein Distance for Map Dissimilarity}\label{sec:sub_wd}

To measure the dissimilarity between $M_1$ and $M_2$, we propose to use the Wasserstein distance, which is a theoretically optimal transport distance between two probability distributions. The $\mathcal{L}_2$ Wasserstein distance between two Gaussian distributions $\mathcal{N}_1$ and $\mathcal{N}_2$ is defined as:
\begin{equation}\label{eq:wasserstein_distance}
\scriptsize
W_2(\mathcal{N}_1, \mathcal{N}_2) = \sqrt{\|\mathbf{\mu}_1 - \mathbf{\mu}_2\|_2^2 + 
\text{tr}\left(\mathbf{\Sigma_1} + \mathbf{\Sigma_2} - 2\left(\mathbf{\Sigma_1}^{1/2} \mathbf{\Sigma_2} \mathbf{\Sigma_1}^{1/2}\right)^{1/2}\right)}.
\end{equation}


The first term, \(\|\mathbf{\mu_1} - \mu_2\|_2^2 \), represents the squared Euclidean distance between the means of the two Gaussian distributions. This term captures the spatial distance between the centers of the distributions, indicating a global shift in the map distribution. 
The second term involves the trace of the covariance matrices, which measures the dissimilarity in the local shape of the distributions.
The covariance matrix describes the shape and orientation of a distribution, particularly the extent of spread along its principal axes.
The square root of the matrix $\mathbf{\Sigma_1}^{1/2} \mathbf{\Sigma_2} \mathbf{\Sigma_1}^{1/2}$ retains the structural characteristics of the distribution, transforming \(\mathbf{\Sigma_2}\) in the spectral space of \(\mathbf{\Sigma_1}\). This transformation reflects the relationship between the shapes and structures of the two distributions. 
If the distributions are similar, the interaction term will be large, resulting in a smaller trace.
Therefore, the Wasserstein distance not only focuses on global differences but also on the shape differences of the distributions, providing a comprehensive measure of dissimilarity, as illustrated in Figure~\ref{fig:waserstain}.
In the context of keyframe selection, a higher Wasserstein distance indicates a significant change in the map distribution. 
This makes it ideal for detecting informative keyframes that enhance overall map completeness. By selecting keyframes based on the Wasserstein distance, we can effectively reduce the map size while preserving essential information and reducing redundancy. However, directly computing the Wasserstein distance for large-scale point cloud maps is computationally expensive due to the need to calculate covariance matrices of the Gaussian components. To achieve real-time performance, we propose a voxel-based Gaussian approximation method, detailed in Section~\ref{subsec:sub_voxel}.
\begin{figure}
    \centering
    \includegraphics[width=0.4\textwidth]{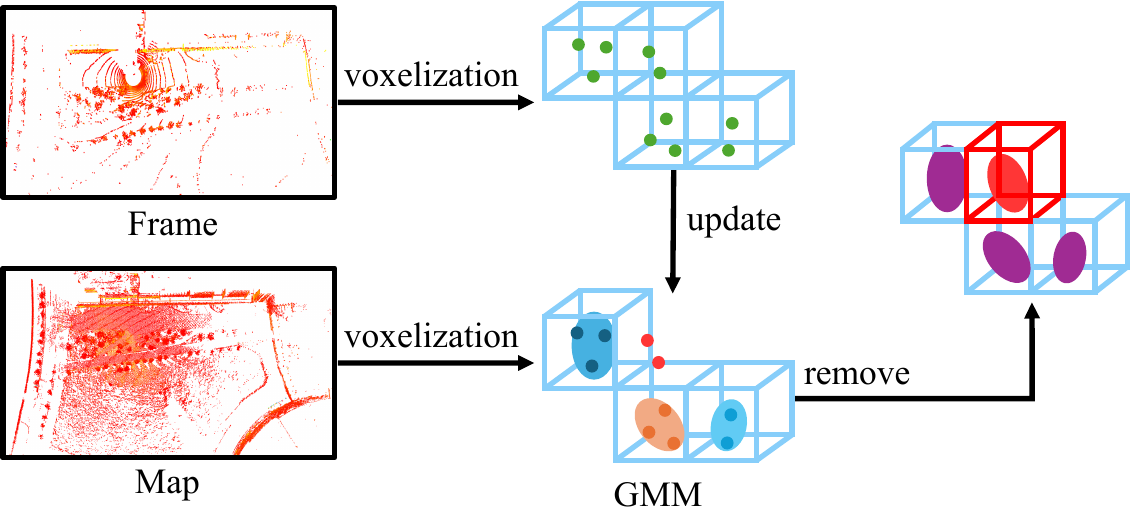}
\caption{
\footnotesize
Incremental update of the GMM map. Each new point cloud frame is transformed into the map based on its pose. The voxels Gaussian parameters are updated point by point, while voxels outside the radius are removed. This process shows the impact of each new frame on the overall map distribution.}
    \label{fig:voxel}
           \vspace{-1.5em}
\end{figure}

\subsection{Voxel Gaussian and Incremental Update}\label{subsec:sub_voxel}

To manage the computational efficiency of a large-scale point cloud map, we divide the local map into voxels of fixed size (e.g., \SI{2.0}{m}) and employ incremental updates for voxels affected by the new point cloud frame $P$. Each voxel is assumed to follow a Gaussian distribution, and the voxels are assumed to be independent of each other. Thus, the map distribution can be approximated as a GMM with each voxel being a Gaussian component (see Fig.~\ref{fig:voxel}).
Let $V_i$ denote the $i$-th voxel, and $\mathcal{N}_i(\mathbf{\mu_i}, \mathbf{\Sigma_i})$ be its corresponding Gaussian distribution. The mean $\mathbf{\mu_i}$ and covariance $\mathbf{\Sigma_i}$ can be computed as:
\begin{equation}
\begin{aligned}
\mathbf{\mu_i} &= \frac{1}{|V_i|} \sum_{x \in V_i} x, \\
\mathbf{\Sigma_i} &= \frac{1}{|V_i| - 1} \sum_{x \in V_i} (x - \mathbf{\mu_i})(x - \mathbf{\mu_i})^\top,
\end{aligned}
\label{eq:gaussian}
\end{equation}
\noindent where $|V_i|$ denotes the number of points in the $i$-th voxel.
When a new point cloud frame $P$ is added to the map, we first transform it into the map frame using the pose $\mathbf{T}$. Then, for each point $p \in P_M$, we compute its corresponding voxel index $(i, j, k)$ :
\begin{equation}\label{eq:hash}
i = \left\lfloor \frac{p_x}{l} \right\rfloor, \quad j = \left\lfloor \frac{p_y}{l} \right\rfloor, \quad k = \left\lfloor \frac{p_z}{l} \right\rfloor,
\end{equation}
\noindent where $l$ is the voxel size and $\lfloor \cdot \rfloor$ denotes the floor function, which rounds its argument down to the nearest integer. The Gaussian distribution of the corresponding voxel can be updated incrementally.
This allows for fast updates and queries of the voxel-based map.

\subsection{Real-time Keyframe Selection}\label{subsec:_real_kf}
Algorithm \ref{alg:keyframe_selection} presents the Wasserstein distance-based keyframe selection process. The algorithm takes a point cloud frame $P$, its estimated pose $\mathbf{T}$, a predefined Wasserstein distance threshold $\tau$, and the voxel size $l$ as inputs. It first initializes a voxel map $M_1$ and computes the Gaussian parameters and point count for each voxel. When a new frame arrives, it is transformed into the map coordinate system and used to update the voxel map incrementally. The average Wasserstein distance between the corresponding voxels in the previous and updated maps is computed using  (\ref{eq:wasserstein_distance}). 
If the average distance exceeds the threshold \(\tau\), the frame is designated as a keyframe. Based on (\ref{eq:wasserstein_distance}), $W_2(\mathcal{N}_1, \mathcal{N}_2)$ is measured in meters, so \(\tau\) can be easily set according to the voxel size. 
Finally, this algorithm processes incoming point cloud frames in real-time, efficiently identifying informative keyframes for large-scale LiDAR mapping.

\begin{algorithm}[t]
\caption{Distribution-aware Keyframe Selection}
\label{alg:keyframe_selection}
\begin{algorithmic}[1]
\State \textbf{Input:} Point cloud frame $P$, pose $\mathbf{T}$, Wasserstein distance threshold $\tau$, voxel size $l$
\State \textbf{Output:} Keyframe decision $b \in {0, 1}$ for frame $P$
\State Initialize voxel map $M_1$
\For{each voxel $V_i$ in $M_1$}
\State Compute and store $\mathbf{\mu_i}$, $\mathbf{\Sigma_i}$, and points count $n_i$ using (\ref{eq:gaussian}), (\ref{eq:hash})
\EndFor
\While{frame $P$ is available}
\State Transform $P$ into the map coordinate $P_M$
\State Initialize voxel map $M_2$
\For{each point $p_M \in P_M$}
\State Incremental update $M_2$.
\EndFor
\State $D_W \gets \textsc{ComputeWassersteinDistance}(M_1, M_2)$
\If{$D_W > \tau$}
\State $b \gets 1$ \Comment{Mark frame $P$ as a keyframe}
\Else
\State $b \gets 0$ \Comment{Mark frame $P$ as a non-keyframe}
\EndIf
\State Remove outside voxels and update map $M_1$ and $M_2$
\EndWhile
\end{algorithmic}
\end{algorithm}

\section{Uncertainty Model for LiDAR SLAM}\label{sec:uncertainty_slam}

Existing approaches\cite{shan2018lego,shan2021liosam} rely on fixed-covariance for pose graph constraints, leading to suboptimal performance and potentially ill-posed optimization in large-scale scenarios. Inspired by our recent work\cite{chen2022fast, hu2024paloc}, this section introduces an uncertainty model to construct a comprehensive LiDAR SLAM system, bridging the gap between theoretical models and practical implementations.

\subsection{Classical LiDAR Measurement Model}\label{subsec:classical_lidar_model}

The measurement models employed in LiDAR SLAM exhibit significant differences from those utilized in visual SLAM due to sensor specifications. Prevalent LiDAR odometry typically base their measurement models on scan-to-map registration, employing techniques such as point-to-plane or point-to-point ICP algorithms. In these models, both scan points and map points are characterized by zero-mean Gaussian noise\cite{xu2022fast,jiao2021robust} or directly neglect\cite{zhang2014loam,shan2021liosam}:
\begin{equation}
\begin{aligned}
        \mathbf{p}_s &= \bar{\mathbf{p}}_s + \mathbf{n}_s, \quad \mathbf{n}_s \sim \mathcal{N}(\mathbf{0}, \mathbf{\Sigma}_s),\\
            \mathbf{p}_m &= \bar{\mathbf{p}}_m + \mathbf{n}_m, \quad \mathbf{n}_m \sim \mathcal{N}(\mathbf{0}, \mathbf{\Sigma}_m),
\end{aligned}
\end{equation}
where $\mathbf{p}_s$ and $\mathbf{p}_m$ denote scan and map points, $\mathbf{n}_s$ and $\mathbf{n}_s$ represent the measurement noise with Gaussian parameters $\mathbf{\Sigma}_s$ and $\mathbf{\Sigma}_m$.

While some approaches\cite{park2017probabilistic, yuan2022efficient} attempt to model noise in mechanical LiDARs by considering factors such as incidence angles and measurement distances, these methods often fall short in complex outdoor environments. Despite showing promise in certain indoor scenarios, they fail to account for intricate environmental noise sources such as varying reflective materials and lighting conditions. Moreover, their computational intensity often renders them less robust, flexible, and efficient compared to the simpler zero-mean Gaussian noise model.
Alternative methods\cite{razlaw2015evaluation} that model point noise based on the Gaussian distribution of the neighborhood points have also been proposed. However, the substantial computational cost associated with neighborhood point searches precludes their application in real-time SLAM systems.
From a first-principles perspective, while it is possible to construct complex noise models, the available sensor data often proves insufficient to support such intricacy. The presence of unmodeled systematic errors or other factors in the data can make precise modeling a formidable challenge. Consequently, the zero-mean Gaussian model for scan points is generally favored for its simplicity, robustness, and practical efficacy.
The general measurement model can be expressed as:
\begin{equation}
    \mathbf{z}_k = h(\mathbf{x}_k, \mathbf{m}) + \mathbf{v}_k.
\end{equation}
where $\mathbf{z}_k$ represents the LiDAR measurement at time $k$, $\mathbf{x}_k$ denotes the robot pose, $\mathbf{m}$ represents the map points, $h(\cdot)$ is the measurement function, and $\mathbf{v}_k \sim \mathcal{N}(\mathbf{0}, \mathbf{\Sigma}_k)$ is the measurement noise.

Despite its limitations, the direct modeling of scan and map points with zero-mean Gaussian noise has gained wide acceptance in practice due to its balanced performance in terms of robustness, accuracy, and simplicity. 
However, the sheer volume of map points renders real-time updates of their covariance $\mathbf{\Sigma}_m$ computationally infeasible. Consequently, the map fails to fully capture the true uncertainty of global map points, leading to an overconfidence in the map during the SLAM process.
This characteristic distinguishes LiDAR odometry from traditional visual odometry in terms of pose covariance behavior. 
Unlike the monotonically increasing pose covariance in world coordinates\cite{carrillo2015monotonicity, rodriguez2018importance}, LiDAR odometry exhibits fluctuating pose covariance magnitudes, as demonstrated in Fig.~\ref{fig:covariance_plots}. Furthermore, the introduction of any mismatched point cloud into the map can precipitate significant odometry drift, underscoring the sensitivity of the system to data quality and matching accuracy.

\begin{figure}
    \centering
    \includegraphics[width=0.4\textwidth]{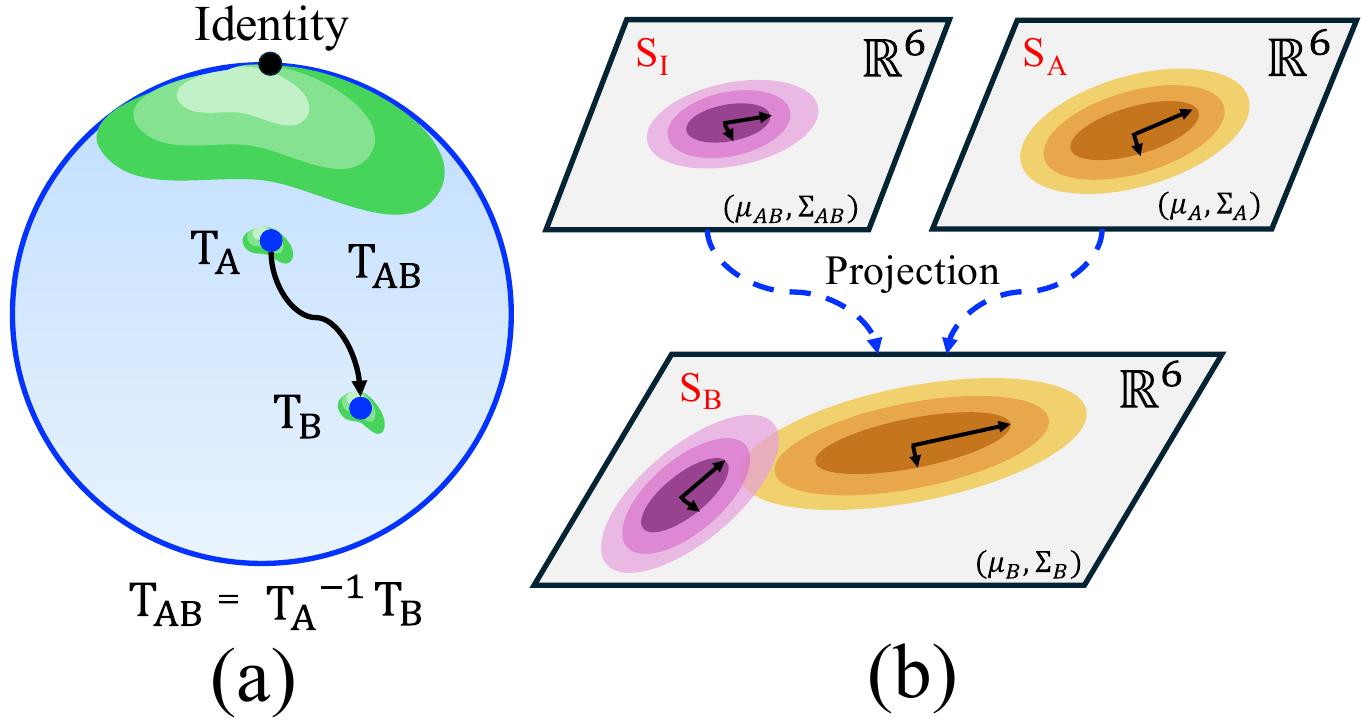}
    \caption{
    \footnotesize 
   Simplified uncertainty propagation with odometry constraints on the $SE(3)$ manifold. \textbf{(a)}: Manifold representation with two consecutive points from $\mathbf{T}_A$ to $\mathbf{T}_B$, illustrating a Gaussian distribution as a 'banana-shaped' green curve. \textbf{(b)}: Odometry constraints derived from $\mathbf{T}_A$ and its uncertainty. Gaussian distributions are combined in the tangent plane at $\mathbf{T}_B$ ($\mathbf{S}_B$) and projected onto the manifold. The Gaussian distribution of the relative pose $\mathbf{T}_{AB}$ is situated in the tangent space ($\mathbf{S}_I$) of its respective manifold , distinct from the origin manifolds of $\mathbf{T}_A$ and $\mathbf{T}_B$.
    }
    \label{fig:manifold}
        \vspace{-1.5em}
\end{figure}

\subsection{Uncertainty Estimation in IEKF}\label{subsec:iekf}

To better understand the uncertainty estimation in odometry, we analyze the Iterated Extended Kalman Filter (IEKF)\cite{he2021kalman}. The IEKF estimates the covariance of the error state, which is defined on the tangent space of the error state product manifold $SO(3) \times \mathbb{R}^3$:
\begin{equation}
    \delta\mathbf{x} = [\delta\boldsymbol{\theta}^\top, \delta\mathbf{p}^\top]^\top,
\end{equation}
where $\delta\boldsymbol{\theta}$  and $\delta\mathbf{p}$ are the error in rotation and position.
The IEKF consists of prediction and update steps.

In the prediction step, IMU measurements (acceleration and angular velocity) are integrated as follows:
\begin{equation}
\begin{aligned}
    \hat{\mathbf{x}}_{k|k-1} &= f(\hat{\mathbf{x}}_{k-1|k-1}, \mathbf{u}_k, \mathbf{w}_k), \\
    \mathbf{P}_{k|k-1} &= \mathbf{F}_k \mathbf{P}_{k-1|k-1} \mathbf{F}_k^\top + \mathbf{G}_k \mathbf{Q}_k \mathbf{G}_k^\top,
\end{aligned}
\end{equation}
where \(f(\cdot)\) is the state transition function, and \(\mathbf{u}_k\) represents the IMU measurements. The process noise \(\mathbf{w}_k\) is set to zero in practice. \(\mathbf{F}_k\) and \(\mathbf{G}_k\) are the Jacobian matrices with respect to the error state and process noise, respectively, and \(\mathbf{Q}_k\) is the process noise covariance matrix.
In the update step, the LiDAR measurement model from Section \ref{subsec:classical_lidar_model} is used:
\begin{equation}
\begin{aligned}
    \mathbf{K}_k &= \mathbf{P}_{k|k-1} \mathbf{H}_k^\top (\mathbf{H}_k \mathbf{P}_{k|k-1} \mathbf{H}_k^\top + \mathbf{R}_k)^{-1},\\
    \hat{\mathbf{x}}_{k|k} &= \hat{\mathbf{x}}_{k|k-1} + \mathbf{K}_k (\mathbf{z}_k - h(\hat{\mathbf{x}}_{k|k-1}, \mathbf{m})),\\
    \mathbf{P}_{k|k} &= (\mathbf{I} - \mathbf{K}_k \mathbf{H}_k) \mathbf{P}_{k|k-1},
\end{aligned}
\end{equation}
where \(\mathbf{H}_k\) is the measurement Jacobian matrix, \(\mathbf{K}_k\) is the Kalman gain, and \(\mathbf{P}_{k|k}\) is the updated error covariance matrix.

For pose graph optimization, we need the covariance of the full state in the world frame represented in $SE(3)$. To convert the error state covariance to the full state covariance, we apply the chain rule:
$\mathbf{\Sigma}_{{SO(3) \times \mathbb{R}}^3} = \mathbf{J} \mathbf{P}_{k|k} \mathbf{J}^\top$,
where $\mathbf{J}$ is the Jacobian of the state w.r.t the error state. When the motion is tiny enough to be infinitesimal, $\mathbf{J}$ can be approximated as an identity matrix.
To transform this covariance to the tangent space of $SE(3)$, we use the adjoint of the pose $\mathbf{T}^w_b$:
$\mathbf{\Sigma}_{SE(3)} = \text{Ad}_{\mathbf{T}^w_b} \mathbf{\Sigma}_{SO(3) \times \mathbb{R}^3} \text{Ad}_{\mathbf{T}^w_b}^\top$,
\noindent where $\text{Ad}_{\mathbf{T}^w_b}$ is the adjoint matrix of $\mathbf{T}^w_b$. This $\mathbf{\Sigma}_{SE(3)}$ is the covariance matrix in the tangent space of $SE(3)$ at the estimated pose that we use for pose graph optimization. It represents the uncertainty of the pose estimate as a Gaussian distribution in the tangent space, which, when projected back to the $SE(3)$ manifold, results in a "banana-shaped distribution" due to the non-Euclidean nature of the space (see Fig.~\ref{fig:manifold}).

\subsection{Uncertainty Modeling in Pose Graph}\label{subsec:pgo}

\subsubsection{Odometry Constraint Covariance}\label{subec:odometry}

We derive the covariance of the relative pose between two consecutive poses, following the approach in\cite{barfoot2014associating, matt2021Reduc,nguyen2018covariance}.
Let $\mathbf{T}_{b_i}^w$ and $\mathbf{T}_{b_{i+1}}^w$ denote the poses at times $i$ and $i+1$, and $\mathbf{\Sigma}_{b_i}$ and $\mathbf{\Sigma}_{b_{i+1}}$ be their corresponding covariance matrices. The relative pose $\mathbf{T}_{b_i b_{i+1}}$ can be expressed as:
\begin{equation}
    \mathbf{T}_{b_i b_{i+1}} = (\mathbf{T}_{b_i}^w)^{-1} \mathbf{T}_{b_{i+1}}^w,
\end{equation}
Assuming that the poses are random variables with Gaussian distributions, we can express them using the exponential map:
\begin{equation}
    \mathbf{T}_{b_i} = \bar{\mathbf{T}}_{b_i} \exp(\mathbf{\epsilon}_{b_i}^{\wedge}), \quad \mathbf{T}_{b_{i+1}} = \bar{\mathbf{T}}_{b_{i+1}} \exp(\mathbf{\epsilon}_{b_{i+1}}^{\wedge}),
\end{equation}
where $\bar{\mathbf{T}}_{b_i}$ and $\bar{\mathbf{T}}_{b_{i+1}}$ are the mean poses, $\mathbf{\epsilon}_{b_i}$ and $\mathbf{\epsilon}_{b_{i+1}}$ are zero-mean Gaussian noises with covariance matrices $\mathbf{\Sigma}_{b_i}$ and $\mathbf{\Sigma}_{b_{i+1}}$.
Using the first-order approximation of the Baker-Campbell-Hausdorff (BCH) formula and the adjoint property of the exponential map, we can derive the covariance of the relative pose $\mathbf{\Sigma}_{b_i b_{i+1}}$:
\begin{equation}
    \mathbf{\Sigma}_{b_i b_{i+1}} \approx \text{Ad}_{\bar{\mathbf{T}}_{b_i b_{i+1}}^{-1}} \mathbf{\Sigma}_{b_i} \text{Ad}_{\bar{\mathbf{T}}_{b_i b_{i+1}}^{-1}}^\top + \mathbf{\Sigma}_{b_{i+1}},
\end{equation}
Further, using the multiplication property of adjoints, we can express this in terms of the individual pose adjoints:
\begin{equation}
     \text{Ad}_{\bar{\mathbf{T}}_{b_i b_{i+1}}^{-1}} = \text{Ad}_{(\mathbf{T}_{b_{i+1}}^w)^{-1}} \text{Ad}_{\mathbf{T}_{b_{i}}^w}.    
\end{equation}
\begin{equation}
        \mathbf{\Sigma}_{b_i b_{i+1}} \approx \text{Ad}_{(\mathbf{T}_{b_{i+1}}^w)^{-1}} \text{Ad}_{\mathbf{T}_{b_{i}}^w} \mathbf{\Sigma}_{b_i} (\text{Ad}_{(\mathbf{T}_{b_{i+1}}^w)^{-1}} \text{Ad}_{\mathbf{T}_{b_{i}}^w})^\top + \mathbf{\Sigma}_{b_{i+1}}.
\end{equation}
It's important to note that this covariance matrix resides in the tangent space of the relative pose at the identity in the $SE(3)$ manifold, as it shown in Fig.~\ref{fig:manifold}. In deriving this covariance, we assume that the two poses are independent which is necessary due to the complexity of computing the joint distribution of poses within the filter. While this assumption approximates the relative pose covariance and may lead to overestimation, as indicated by\cite{mangelson2020characterizing, wu2022quadratic}, it can be considered a small perturbation for short-term motion. In our real-world experiments in Section~\ref{sec:module_exp}, we found that this assumption works well\cite{hu2024paloc}.

\subsubsection{Loop Constraint Covariance}\label{subsec:prior_map_constraint}

As formulated in Section \ref{subsec:prior_map_assisted_pgo}, we treat the multi-session SLAM problem as a Pose-SLAM problem. The constraints between the current pose and the nearest historical pose, which is from the combined map of old and new sessions, can be modeled as special loop closure constraints. 
Unlike the odometry covariance, the loop closure constraint is an independent observation. This is a key difference in how we handle odometry and loop closure uncertainty in the pose graph.
Unlike traditional low-frequency loop closures, these constraints are applied to each frame, although their mathematical formulation remains consistent. When a match is found between the current pose $\mathbf{T}_{c}^w$ and a historical pose $\mathbf{T}_{p}^w$, we estimate the relative pose $\mathbf{T}_{pmc}$ and its covariance $\mathbf{\Sigma}_{pmc}$ using a point-to-plane ICP algorithm:
\begin{equation}
    \mathbf{T}_{pmc} = (\mathbf{T}_{p}^w)^{-1} \mathbf{T}_{c}^w,
\end{equation}
The covariance $\mathbf{\Sigma}_{pmc}$ is approximated using the inverse of the Hessian matrix computed during the ICP process:
$\mathbf{\Sigma}_{pmc} \approx (\mathbf{J}^\top \mathbf{W} \mathbf{J})^{-1}$,
where $\mathbf{J}$ is the Jacobian matrix of the ICP, and $\mathbf{W}$ is the noise matrix for the point correspondences.

\begin{figure*}
    \centering
    \includegraphics[width=0.8\textwidth]{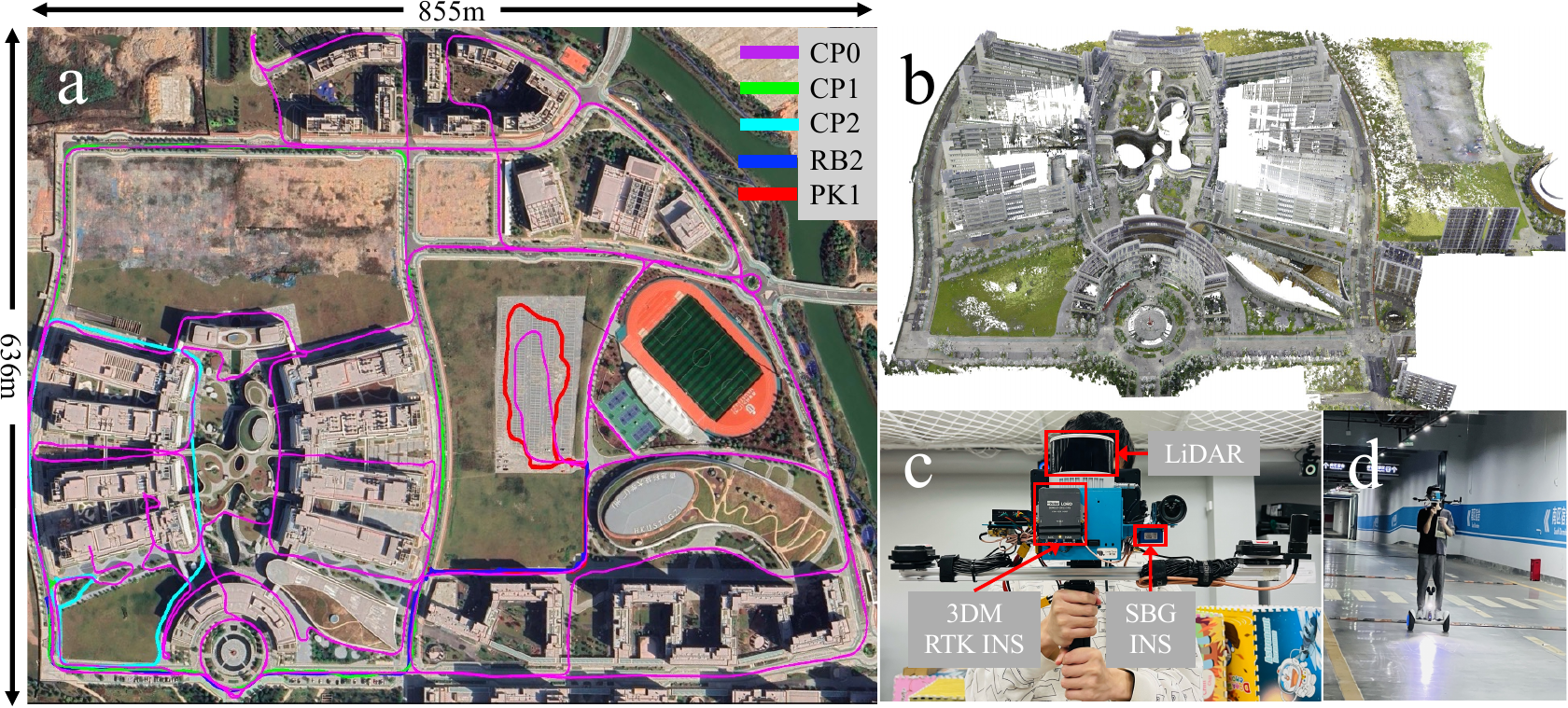}
    \caption{
    \footnotesize
    \textbf{Experimental setup}. (a) Trajectories of the MS-dataset projected onto Google Maps, covering the majority of the campus area. (b) Ultra-high resolution ground truth point cloud map\cite{wei2024fusionportablev2}. (c) Handheld multi-sensor platform. (d) Self-balancing vehicle employed for data collection. }
    \label{fig:sensor_kit}
           \vspace{-1.5em}
\end{figure*}

\subsubsection{Prior Factor Covariance}\label{subsec:prior_factor_covariance}

Although there is only one prior constraint at the starting point of the pose graph, it plays a crucial role in handling the Gauge Freedom problem\cite{zhang2018comparison}, which could otherwise lead to ill-posed situations in graph optimization.
In our approach, we directly use the pose covariance provided by the odometry front-end (Section~\ref{sec:uncertainty_slam}) for the prior factor noise. This unifies the scale of pose covariance across the system and avoids ill-posed optimization problems.
The prior factor can be formulated as:
\begin{equation}
    \mathbf{e}_{pr} = \log((\mathbf{T}_0^w)^{-1} \mathbf{T}_1^w)^{\vee},
\end{equation}
where $\mathbf{T}_0^w$ is the initial pose estimate, $\mathbf{T}_1^w$ is the first pose in the graph, and $(\cdot)^{\vee}$ is the vee operator that maps an element of $SE(3)$ to its vector representation.
The covariance of this prior factor is set to the initial pose covariance from the odometry front-end:
$\mathbf{\Sigma}_{pr} = \mathbf{\Sigma}_{SE(3)}|_{t=0}$.
By using this consistent covariance scaling, we ensure numerical stability in the pose graph optimization while maintaining the uncertainty information from the front-end estimation.

\section{Dataset and Experiments Setup}\label{sec:dataset}

\subsection{Datasets and Setup}\label{subsec:dataset}
We implement the MS-Mapping algorithm using C++, ROS Noetic, GTSAM \cite{dellaert2012factor}, and Open3D \cite{zhou2018open3d}. Experiments are conducted on a personal desktop computer with an Intel Core $i7$-$12700$K CPU, \SI{96}{GB} of RAM, and a \SI{2}{TB} hard drive. To evaluate the proposed MS-Mapping system, we employ several real-world public datasets, with sensor specifications summarized in Table~\ref{tab:dataset_characteristics}. 
It should be noted that datasets providing high-precision dense ground truth point cloud maps are rare and typically collected in small-scale environments, which are insufficient for large-scale map evaluation.
Our recent work, FusionPortableV2 \cite{wei2024fusionportablev2}, addresses this limitation with an ultra-high-resolution \SI{560}{m}$\times$\SI{640}{m} RGB point cloud map, encompassing various indoor and outdoor scenes. The ground truth map, collected by Leica RTC$360$ (outdoor) and BLK$360$ (indoor), boasts an average point accuracy of \SI{3.425}{mm}, as shown in Fig.~\ref{fig:sensor_kit} (b). This enables comprehensive data collection for indoor and outdoor experiments within the coverage of the ground truth map.

\begin{table}[t]
\centering
\caption{Characteristics of Data Sequences in the MS-Dataset}
\label{tab:ms_dataset}
\renewcommand{\arraystretch}{1.0}
    \label{tab:ms_characteristics}
\begin{threeparttable}
\begin{tabular}{lp{4cm}cc}
    \toprule[0.03cm]
    \textbf{Sequence} & \textbf{Environment} & \textbf{Dur.} & \textbf{Dist.} \\
    \midrule
    \texttt{CC1} & Campus with many loops  & 1736 & 4.79 \\
    \texttt{CS1} & Indoor-outdoor transition  & 614 & 0.72 \\
    \texttt{CP0} & Scenarios with many loops  & 6413 & 10.8 \\
    \texttt{CP1} & Outdoor buildings & 748 & 1.74 \\
    \texttt{CP2} & Road and buildings   & 499 & 1.75 \\
    \texttt{CP3} & Outdoor small-scale  & 240 & 0.15 \\
    \texttt{CP5} & Building with large loop  & 1084 & 2.58 \\
    \texttt{PK1} & Degenerate parkinglot  & 502 & 0.97 \\
    \texttt{RB2} & Road with two loops & 822 & 1.06 \\
    \texttt{RB3} & Road and buildings  & 636 & 1.41 \\
    \texttt{RB4} & Campus road  & 560 & 1.09 \\
    \texttt{IA3} & Building and indoor  & 775 & 0.72 \\
    \texttt{IA4} & Building  & 585 & 1.02 \\
    \texttt{NG} & Outdoor  & 1798 & 4.50 \\

    \bottomrule[0.03cm]
\end{tabular}
        \begin{tablenotes}[flushleft]
            \footnotesize
            \item \textbf{Note:} \textbf{Dist.}: sequence distance (\SI{}{km}). \textbf{Dur.}: sequence duration (\SI{}{s}).
        \end{tablenotes}
\end{threeparttable}
    \vspace{-1.5em}
\end{table}

For the multi-session LiDAR mapping task, we design a handheld hardware platform to collect diverse data sequences. This dataset, named MS-Dataset, would be made publicly available.
The sensor suite consists of a Pandar XT$32$ LiDAR with a measurement noise of approximately $\pm$\SI{1}{cm} within a \SI{40}{m} range, an SBG-INS with a \SI{100}{Hz} IMU, and a MicroStrain $3$DM-GQ$7$ RTK-INS with a \SI{700}{Hz} IMU. We utilize a \SI{30}{Hz} 6-DoF trajectory provided by the $3$DM-GQ$7$ for  localization evaluation. Fig.~\ref{fig:sensor_kit}(c) and Fig.~\ref{fig:sensor_kit}(d) showcases our handheld sensor acquisition platform.
Fig.~\ref{fig:sensor_kit}(a) displays the five collected data sequence projected onto Google Map. The magenta \texttt{CP0} dataset path, collected using a mini vehicle, covers a distance exceeding \SI{10.8}{km}. The \texttt{CP1}, \texttt{CP2}, \texttt{RB2}, and \texttt{PK1} sequences are collected by a self-balancing vehicle or handheld setup, as shown in Fig.~\ref{fig:sensor_kit}(d). The different sensor configurations and platforms help us better assess the algorithm under various hardware and motion characteristics. The related sequences in the MS-Dataset are presented in Table~\ref{tab:ms_characteristics}.
Other typical public dataset used in our experiments are characterized as follows:
\begin{enumerate}
    \item \textbf{Newer College}: This dataset includes various mobile mapping sensors carried by hand at typical walking speeds through New College, Oxford, covering nearly \SI{6.7}{km} and encompassing built environments, open spaces, and vegetated areas.
    \item \textbf{FusionPortable}: This dataset features longer indoor sequences collected with handheld devices.
    \item \textbf{FusionPortableV2}: A multi-platform and scalable dataset, ranging from indoor to large-scale outdoor scenes.
    \item \textbf{Urban-Nav}: A large-scale urban dataset collected by cars.
\end{enumerate}

\subsection{Evaluation Metrics}\label{subsec:experi_metric}
\subsubsection{Localization}
To evaluate the localization performance of the estimated trajectories, we employ the Absolute Trajectory Error (ATE), as implemented in the EVO library\cite{grupp2017evo}, which is defined as:
$\textbf{ATE} = \sqrt{\frac{1}{n} \sum_{i=1}^{n} \lVert \mathbf{t}_i - \mathbf{\hat{t}}_i \rVert^2}$,
where $\mathbf{t}_i$ and $\mathbf{\hat{t}}_i$ denote the ground truth and estimated positions at time step $i$, and $n$ is the number of poses in the trajectory. 

\begin{table}[t]
    \centering
    \caption{Dataset Characteristics with Sensor Configurations}
    \label{tab:dataset_characteristics}
    \renewcommand{\arraystretch}{1.0}
    \begin{threeparttable}
        \begin{tabular}{@{}p{2cm}llcc@{}}
            \toprule[0.03cm]
            \textbf{Dataset} & \textbf{LiDAR} & \textbf{IMU} & \textbf{Dist.} & \textbf{Map} \\
            \midrule[0.03cm]
            MS-Dataset & Pandar XT$32$ & SBG/$3$DM & 27.54 & 0.3 \\
            Newer College & Ouster-$64$/$128$ & Integrated & 6.7 & 0.6 \\
            UrbanNav & Velodyne-$32$ & Xsense & $\geq$29.4 & -- \\
            FusionPortable & Ouster-$128$ & STIM$300$ & 3.0 & 0.6 \\
            FusionPortableV2 & Ouster-$128$ & STIM$300$/$3$DM & 38.7 & 0.3 \\
            \bottomrule[0.03cm]
        \end{tabular}
        \begin{tablenotes}[flushleft]
            \footnotesize
            \item \textbf{Note:}  \textbf{Dist.}: dataset distance (\SI{}{km}). \textbf{Dur.}: sequence duration (\SI{}{s}). \textbf{Map}: ground truth accuracy (\SI{}{cm}). \textbf{--} indicates ground truth is unavailable.
        \end{tablenotes}
    \end{threeparttable}
    \vspace{-1.5em}
\end{table}

\subsubsection{Mapping}
To assess the map accuracy, we utilize the Accuracy (AC) and Chamfer Distance (CD)\cite{hu2024paloc}. During the registration process, we set the KNN search distance to \SI{1.0}{m} and consider point pairs with a distance smaller than \SI{0.5}{m} as inliers. The Map Accuracy calculates the average RMSE of the inlier point pairs, defined as:
\begin{equation*}
\textbf{AC} = \sqrt{\frac{1}{|\mathcal{I}|} \sum_{(\mathbf{p}, \mathbf{q}) \in \mathcal{I}} \lVert \mathbf{p} - \mathbf{q} \rVert^2},
\end{equation*}
where $\mathcal{I}$ represents the set of inlier point pairs, and $(\mathbf{p}, \mathbf{q})$ denotes a pair of corresponding points from the estimated and ground truth maps.
The Chamfer Distance directly measures the distance between point clouds, computed as:
\begin{equation*}
\textbf{CD} = \frac{1}{|\mathcal{P}|}\sum_{\mathbf{p} \in \mathcal{P}} min_{\mathbf{q} \in \mathcal{Q}} \lVert \mathbf{p} - \mathbf{q} \rVert_2 + \frac{1}{|\mathcal{Q}|}\sum_{\mathbf{q} \in \mathcal{Q}} min_{\mathbf{p} \in \mathcal{P}} \lVert \mathbf{p} - \mathbf{q} \rVert_2,
\end{equation*}
where $\mathcal{P}$ and $\mathcal{Q}$ represent the sets of points in the estimated and ground truth maps. CD measures the overall distance difference between two point cloud maps, while AC finely characterizes the local distance between them.

Furthermore, we utilize the Mean Map Entropy (MME) to evaluate the local consistency of the generated maps\cite{droeschel2018efficient}. MME quantifies the local compactness of points in the map by calculating the average entropy within a local radius $r$ around each map point $q_k$. The entropy $h$ for a map point $p_k$ is calculated by:
$h(q_k) = \frac{1}{2} \ln |2\pi e \mathbf{\Sigma(p_k)}|$,
where $\mathbf{\Sigma(p_k)}$ is the covariance of mapped points in a local radius $r$ around $p_k$. We select $r = \SI{0.1}{m}$ (indoor) or $r = \SI{0.2}{m}$ (outdoor)  with at least 10 points. The MME is averaged over all map points:
$\textbf{MME} = \frac{1}{Q} \sum_{k=1}^{Q} h(p_k)$,
where $Q$ is the total number of map points. A lower MME indicates higher point density and better map accuracy and consistency.

\begin{figure}
    \centering
    \includegraphics[width=0.4\textwidth]{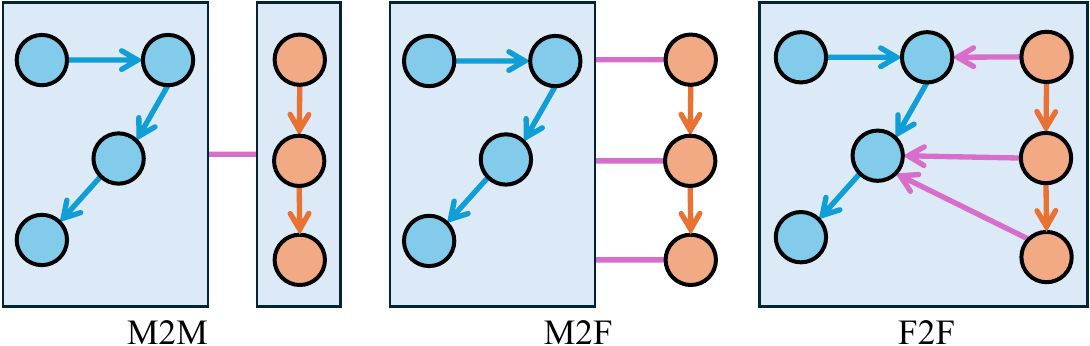}
    \caption{
    \footnotesize
    Principles of map merging for baseline algorithms. \textbf{M2M}: Direct registration of the new map onto the old map. \textbf{M2F}: Localization of the new session point cloud frame-by-frame on the old map to establish prior pose constraints. \textbf{F2F}: Localization of the new session point cloud frames on the combined (new + old) map to create loop closure constraints.
    }
    \label{fig:baseline_compare}
           \vspace{-1.5em}
\end{figure}

\subsection{Baseline}\label{subsec:experi_baseline}

For the evaluation of multi-session systems, it is crucial to acknowledge that the performance of different odometry algorithms varies, and factors such as point cloud sampling size and other parameters can significantly impact the system performance.
This paper primarily focuses on back-end optimization; thus, direct comparisons with systems specializing in place recognition\cite{xu2023ring}, multi-robot real-time exploration, or those using different or improved LiDAR odometry\cite{yuan2023lta} are not feasible. Such comparisons would be unfair and would not accurately reflect or validate the system performance. However, place recognition modules can be directly integrated into our system.
Fig.~\ref{fig:baseline_compare} illustrates the principles of three types of multi-session mapping algorithms. 
\begin{enumerate}
    \item \textbf{M2M} considers both independent maps as a unity, directly registering the new map onto the old map to achieve map merging. This approach, commonly used in surveying, assumes that both point cloud maps are dense, highly accurate, and have a large overlapping area (typically larger than 60\%). We use CloudCompare\footnote{\url{https://www.danielgm.net/cc/}} to register the two dense point cloud maps.
    \item \textbf{M2F} centers on the pose graph, using each frame of the new session to localize on the old map and build prior pose constraints. This extends the pose graph and achieves map merging through Pose-SLAM. This principle is often applied in real-time localization algorithms based on prior maps and assumes that the old map is more reliable than the new map (typically by an order of magnitude in accuracy). We implement this approach by referencing FAST\_LIO\_LOCALIZATION\footnote{\url{https://github.com/HViktorTsoi/FAST\_LIO\_LOCALIZATION}}.
    \item \textbf{F2F}, based on the pose graph, utilizes each frame in new session to sequentially registration with local maps. This method constructs loop closure constraints to extend the pose graph and builds the merged map through graph optimization. It assumes minimal geometric structure changes in the overlapping regions.  We implement this by referencing FAST\_LIO\_SAM\footnote{\url{https://github.com/engcang/FAST-LIO-SAM}}. MS-Mapping (\textbf{MS}) also follows this approach. 
\end{enumerate}
For these baseline algorithms, we guarantee that the odometry and pose graph parameters for M2F and F2F are identical to the  proposed algorithm. This includes the sampling size of the point clouds and the threshold settings for loop closure. We set the prior rotational and translational noise for M2F and F2F to 0.01 and 1.0, and the noise for loop closure detection to 0.1. All three algorithms are implemented within the same framework for fair comparison.

\begin{table}[t]
\centering
\caption{Comparison of PGO Noise Parameters}
\label{tab:algorithm-params-pgo}
\renewcommand{\arraystretch}{1.0}
\begin{threeparttable}
\begin{tabular}{@{}llcc@{}}
\toprule[0.03cm]
\textbf{Factor} & \textbf{Algorithm} & \textbf{Translation} & \textbf{Rotation} \\
\midrule[0.03cm]
\multirow{2}{*}{Prior} & \textbf{M2F/F2F} & 1e0 & 1e-2 \\
                        & \textbf{MS} & \multicolumn{2}{c}{Unified noise scale: 1e-2} \\
\midrule
\multirow{2}{*}{Odom}   & \textbf{M2F/F2F} & 1e-6 & 1e-8 \\
                        & \textbf{MS} & \multicolumn{2}{c}{Unified noise scale: 1e-2} \\
\midrule
\multirow{2}{*}{Loop Closure}   & \textbf{M2F/F2F} & 1e-1 & 1e-1 \\
                        & MS & \multicolumn{2}{c}{Unified noise scale: 1e-2} \\

\bottomrule[0.03cm]
\end{tabular}
\end{threeparttable}
\vspace{-1.5em}
\end{table}

\subsection{Pose Graph Parameters Tuning}\label{sub_sec:pgo_para_tuning}

A typical pose graph optimization backend generally includes three types of factors, each requiring manual settings for the covariance matrix, especially for translation and rotation noise, resulting in a Gaussian diagonal noise model. Table~\ref{tab:algorithm-params-pgo} presents the parameter settings for F2F, M2F, and our proposed MS algorithm. Both M2F and F2F use a set of typical noise parameters\cite{shan2021liosam}, which require six parameters to be configured. An inappropriate setting for prior noise parameters can directly cause the Gauge Freedom problem\cite{zhang2018comparison}, while improper loop closure parameters can lead to the 'pose graph relaxation' issue (Section~\ref{sub_sec:chall_acc} and Section~\ref{subsec:pgo} ), both resulting in ill-posed graph optimization. Particularly in large-scale scenarios, avoiding ill-posed optimization necessitates scene-specific parameters tuning, which cannot fundamentally solve the problem.

In contrast, the MS algorithm only requires a single noise level parameter, typically set to 0.01 for indoor scene and 100 for outdoor environments, as determined by the point-to-plane noise parameters of loop closure detection (\ref{eq:loop_cov}). When using the point-to-plane registration to calculate loop closure constraints, its accuracy is generally lower than odometry relative pose. This is because odometry poses are derived from a highly precise initial pose over a short time, whereas the initial value for loop closure is usually less accurate, resulting in slower convergence and lower pose accuracy. Additionally, when calculating the covariance for loop closure relative pose, it is essential to consider that the point-to-plane measurement noise may not match that of odometry. Therefore, we only use a noise scale parameter to adapt to different indoor and outdoor scenarios, show its flexibility and adaptability of our MS-Mapping system.

\begin{figure*}
    \centering
    \includegraphics[width=0.9\textwidth]{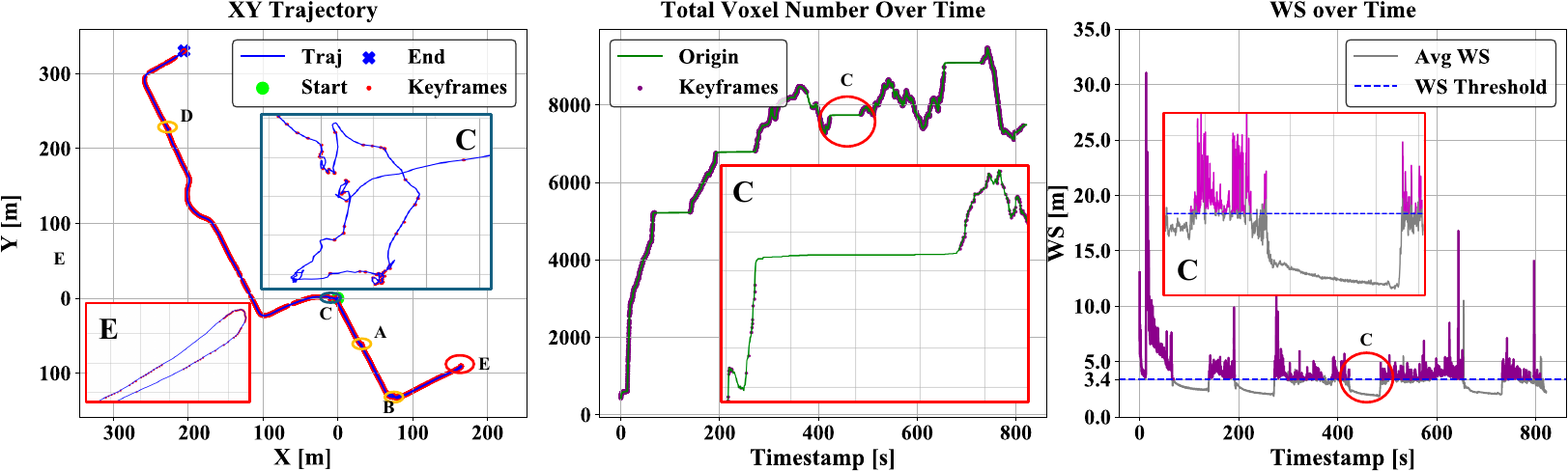}
    \caption{
    \footnotesize
    Keyframe selection and temporal analysis on the \texttt{RB2} dataset. \textbf{Left:} Trajectory with selected keyframes (red). \textbf{Middle:} Temporal voxel count, keyframes marked in brown. \textbf{Right:} Per-frame average Wasserstein distance, keyframes exceeding blue threshold line.}
    \label{fig:keyframe_case}
          \vspace{-1.5em}
\end{figure*}

\section{Modularized Experimental  Results}\label{sec:module_exp}
\subsection{Keyframe Selection Experiments}

\subsubsection{Experiment Design}
\begin{figure}
    \centering
    \includegraphics[width=0.45\textwidth]{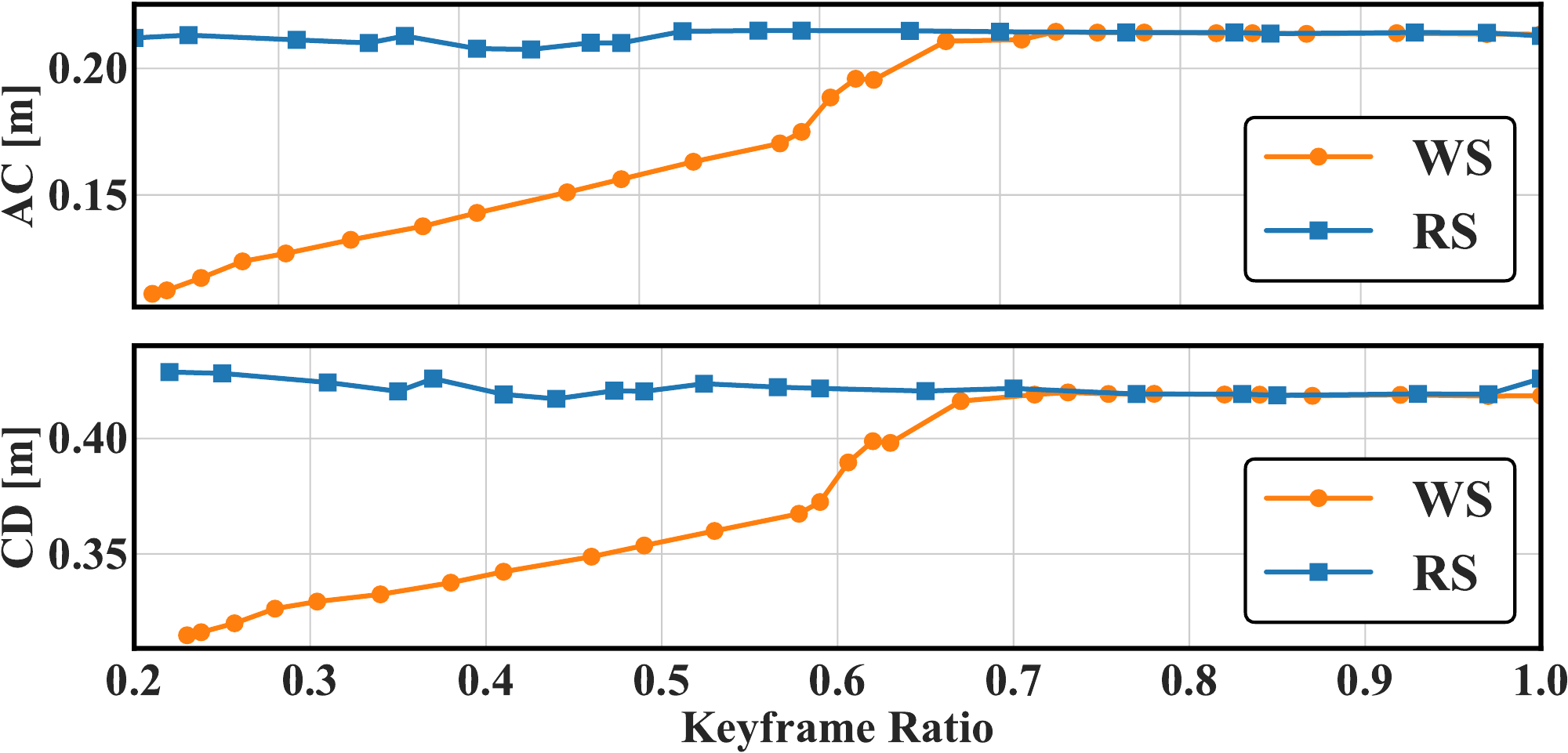}
\caption{
\footnotesize
Map accuracy comparison between the proposed and radius-based methods on the \texttt{RB2} dataset across varying keyframe ratios.}
    \label{fig:keyframe}
          \vspace{-1.5em}
\end{figure}

We evaluate the efficiency of our proposed distribution-aware keyframe selection method based on the Wasserstein distance (WS) against the widely used radius-based (RS) method \cite{shan2021liosam,shan2018lego,kim2022lt}. The RS approach selects keyframes when motion state changes exceed predefined thresholds (e.g., \SI{0.1}{m} for translation or \SI{0.1}{rad} for rotation), a common technique in SLAM systems.
We do not specifically evaluate the efficiency improvement of graph optimization due to keyframe selection, as a reduced number of keyframes naturally decreases the pose graph size, thereby accelerating optimization without further proof. The proposed WS method calculates the overall difference of each point cloud frame relative to the map, requiring additional computation time. Although the WS method achieves real-time filtering, it is inherently slower than the RS method.
This performance difference is evident in the timing analysis presented in Section~\ref{sub_sec:runtime}.
Therefore, our experimental design includes two main components:
\begin{itemize}
    \item Comparative analysis of varying keyframe ratios (KFR) on mapping accuracy (Fig.~\ref{fig:keyframe}).
    \item Temporal analysis of  voxel count and Wasserstein distance for the proposed method with a typical case (Fig.~\ref{fig:keyframe_case}).
\end{itemize}
These comprehensive analysis provides insights into the strengths and limitations of our new method. 
For consistency, we maintained a voxel size of \SI{5.0}{m} and a map radius of \SI{800}{m} for the WS method across all experiments. 

\subsubsection{Comparison Results}

We first compared the map accuracy and Chamfer Distance of our proposed WS method against the RS method across different KFRs on the \texttt{RB2} sequence. Given the differing principles of the WS and RS methods, achieving identical KFRs is challenging. Therefore, we adjusted their respective thresholds to ensure comparable KFRs, ensuring a fair comparison.
By examining the mapping accuracy of both methods, we obtained various KFRs, resulting in over 45 experimental configurations. 
Figure~\ref{fig:keyframe} presents the results, showing that the WS method consistently outperforms the RS method across the entire KFR range. This indicates that our method not only maintains mapping precision but also improves it.
The Chamfer Distance, which quantifies the overall similarity between the original and updated maps, further validates the effectiveness of our approach in preserving essential information. Notably, when the KFR approaches 0.7, both AC and CD for the WS method exhibit noticeable convergence. This suggests that at this KFR level, the selected keyframes comprehensively describe the entire scene. Exceeding a KFR of 0.7 may lead to data redundancy.
Considering that most LiDAR sensors operate at \SI{10}{Hz}, additional point clouds do not provide extra map information at low moving speeds, leading to redundant data. These findings highlight the importance of considering map accuracy when evaluating keyframe selection techniques in SLAM systems. 
The WS method offers a promising approach for keyframe selection, maintaining excellent map accuracy and reducing data redundancy across various KFR.

\begin{table}[t]
    \centering
    \caption{Quantitative Comparison of UPGO and FPGO}
    \label{tab:comparison_ate}
    \renewcommand{\arraystretch}{1.0}
    \setlength{\tabcolsep}{4pt}
    \begin{threeparttable}
        \begin{tabular}{@{}l c c c c c@{}}
            \toprule[0.03cm]
            \multirow{2}{*}{\textbf{Sequence}} & \textbf{Dist.} & \textbf{Dur.} & \multicolumn{2}{c}{\textbf{ATE [\si{\centi\meter}] $\downarrow$}} & \textbf{Improv.} \\
            \cmidrule(lr){2-2} \cmidrule(lr){3-3} \cmidrule(lr){4-5}
            & \textbf{[\SI{}{km}]} & \textbf{[\SI{}{s}]} & \textbf{FPGO} & \textbf{UPGO} & [\%] \\
            \midrule[0.03cm]
            \multicolumn{6}{@{}l}{\textbf{MS-Dataset}} \\
            \midrule
            \texttt{GC4}     & 0.46   & 365 & 55.53  & \textbf{55.16} & 0.67 \\
            \texttt{GC4*}    & 0.46   & 365 & 66.09  & \textbf{66.03} & 0.09 \\
            \texttt{PK1}     & 0.97   & 502 & 23.25  & \textbf{22.68} & 2.45 \\
            \texttt{PK1*}    & 0.97   & 502 & 18.98  & \textbf{18.32} & 3.48 \\
            \texttt{RB2}     & 1.06   & 822 & 20.74  & \textbf{18.50} & 10.80 \\
            \texttt{RB2*}    & 1.06   & 822 & 24.48  & \textbf{22.42} & 8.41 \\
            \texttt{RB3}     & 1.41   & 636 & 96.62  & \textbf{88.42} & 8.49 \\
            \texttt{RB3*}    & 1.41   & 636 & 94.48  & \textbf{92.85} & 1.73 \\
            \texttt{CP1*}    & 1.74   & 748 & 245.64 & \textbf{148.09} & 39.71 \\
            \texttt{CP5}     & 2.49   & 1084 & 121.42 & \textbf{114.91} & 5.36 \\
            \texttt{CP5*}    & 2.49   & 1084 & 119.79 & \textbf{113.83} & 4.97 \\
            \texttt{CC1}     & 4.79   & 1736 & {$\times$} & \textbf{110.21} & -- \\
            \texttt{CC1*}    & 4.79   & 1736 & {$\times$} & \textbf{93.99} & -- \\
            \midrule
            \multicolumn{6}{@{}l}{\textbf{Newer College}} \\
            \midrule
            \texttt{cloister-seg1} & 0.14 & 92 & 6.77  & \textbf{5.71} & 15.66 \\
            \texttt{stairs} & 0.06 & 119 & 10.74  & \textbf{9.85} & 8.29 \\
            \texttt{parkland-seg0} & 0.23 & 157 & 14.03 & \textbf{13.93} & 0.71 \\
            \texttt{math-medium} & 0.28 & 177 & \textbf{13.39} & \textbf{13.39} & 0.00 \\
            \texttt{cloister-seg0} & 0.29 & 186 & 8.82  & \textbf{8.13} & 7.82 \\
            \texttt{quad-hard} & 0.23 & 188 & 6.98  & \textbf{4.56} & 34.67 \\
            \texttt{quad-medium} & 0.26 & 191 & 6.05  & \textbf{6.02} & 0.50 \\
            \texttt{quad-easy} & 0.25 & 199 & 7.20  & \textbf{7.07} & 1.81 \\
            \texttt{parkland-seg1} & 0.19 & 204 & 11.00 & \textbf{10.30} & 6.36 \\
            \texttt{math-easy} & 0.25 & 215 & 10.63 & \textbf{9.75} & 8.29 \\
            \texttt{math-hard} & 0.32 & 244 & 7.19 & \textbf{6.95} & 3.34 \\
            \texttt{cloister}      & 0.43 & 278 & 14.07 & \textbf{13.14} & 6.61 \\
            \texttt{mount}         & 0.65 & 500 & 17.94 & \textbf{17.79} & 0.84 \\
            \texttt{short-exp10$\dagger$}   & 0.51 & 528 & 36.58     & \textbf{36.57}     & 0.27 \\
            \texttt{parkland0}     & 1.20 & 769 & 28.24 & \textbf{26.25} & 6.55 \\
            \texttt{long-exp1}    & 0.85 & 835 & 32.89     & \textbf{32.43}     & 1.40 \\
            \texttt{short-exp11}   & 0.82 & 862 & 46.15     & \textbf{45.05}     & 2.38 \\
            \texttt{long-exp0}    & 0.92 & 1002 & 25.52 & \textbf{24.99} & 2.08 \\
            \texttt{short-exp0}   & 0.91 & 1002 & 46.61     & \textbf{46.11}     & 1.07 \\

            \midrule
            \multicolumn{6}{@{}l}{\textbf{FusionPortable}} \\
            \midrule
            \texttt{canteen\_day}   & 0.26 & 267 & 6.96  & \textbf{6.93} & 0.43 \\
            \texttt{garden\_day}   & 0.29 & 288 & 4.58  & \textbf{4.55} & 0.66 \\
            \texttt{corridor\_day} & 0.66 & 572 & 31.12 & \textbf{29.77} & 4.34 \\
            \texttt{escalator\_day} & 0.59 & 714 & 18.32 & \textbf{18.30} & 0.11 \\
            \texttt{building\_day} & 0.68 & 599 & 17.02 & \textbf{16.44} & 3.41 \\

            \midrule
            \multicolumn{6}{@{}l}{\textbf{Urban-Nav}} \\
            \midrule
            \texttt{TST}     & 3.64 & 785 & 308.81 & \textbf{252.11} & 18.36 \\
            \texttt{Whampoa} & 4.51 & 1536 & 617.23 & \textbf{465.74} & 24.54 \\
            \texttt{Mongkok} & 4.86 & 1791 & {$\times$} & \textbf{292.90} & -- \\

            \bottomrule[0.03cm]
        \end{tabular}
        \begin{tablenotes}[flushleft]
            \footnotesize
            \item \textbf{Note}: Bold values indicate better performance. $\times$ indicates ill-posed optimization. *: Sequences using 3DM IMU. $\dagger$: No loop closure. \textbf{Improv.}: Relative improvement of UPGO over FPGO.
        \end{tablenotes}
    \end{threeparttable}
    \label{tab:traj_ape_rpe_half}
    \vspace{-2em}
\end{table}

\subsubsection{Temporal Analysis}


To validate the effectiveness of the WS method in real-world scenarios, we tested it on the \texttt{RB2} sequence. The hardware suit was placed on the ground at four specified locations, remaining completely stationary for several seconds (regions A, B, C, and D in Fig.~\ref{fig:keyframe_case}). During this period, there were no changes in the LiDAR point clouds, contributing minimally to the environmental map. We observed the Wasserstein distance values to demonstrate the effectiveness of our new method. 
In the left subfigure of Fig.~\ref{fig:keyframe_case}, the enlarged region C shows keyframes in red, with a KFR of 47.5\%. In this stationary region, our predefined threshold filtered out a large number of keyframes. 
Additionally, we constructed and incrementally updated a GMM map for keyframe selection, setting the map range to \SI{800}{m} to fully cover the \texttt{RB2} data. The middle panel of Fig.~\ref{fig:keyframe_case} illustrates this process. The voxel count increased rapidly during the construction of the GMM map and stabilized once the map reached a certain extent. In stationary regions such as region C, the voxel count remained constant, confirming the theory presented in Section~\ref{sec:ws_keyframe}.
In the keyframe selection process, the right panel of Fig.~\ref{fig:keyframe_case} shows the average Wasserstein distance change over time. Notably, in the stationary region C, the values were significantly lower than in non-stationary regions (gray segments), further validating the capability of our new method. 
The average Wasserstein distance in stationary regions, combined with voxel size, also help us set the threshold for keyframe selection using the WS method.
This experiment demonstrates that the proposed WS method effectively captures and quantifies the contribution of keyframes to the map based on the scenes, aiding in the better selection of keyframes.

 \begin{figure}
    \centering
    \includegraphics[width=0.45\textwidth]{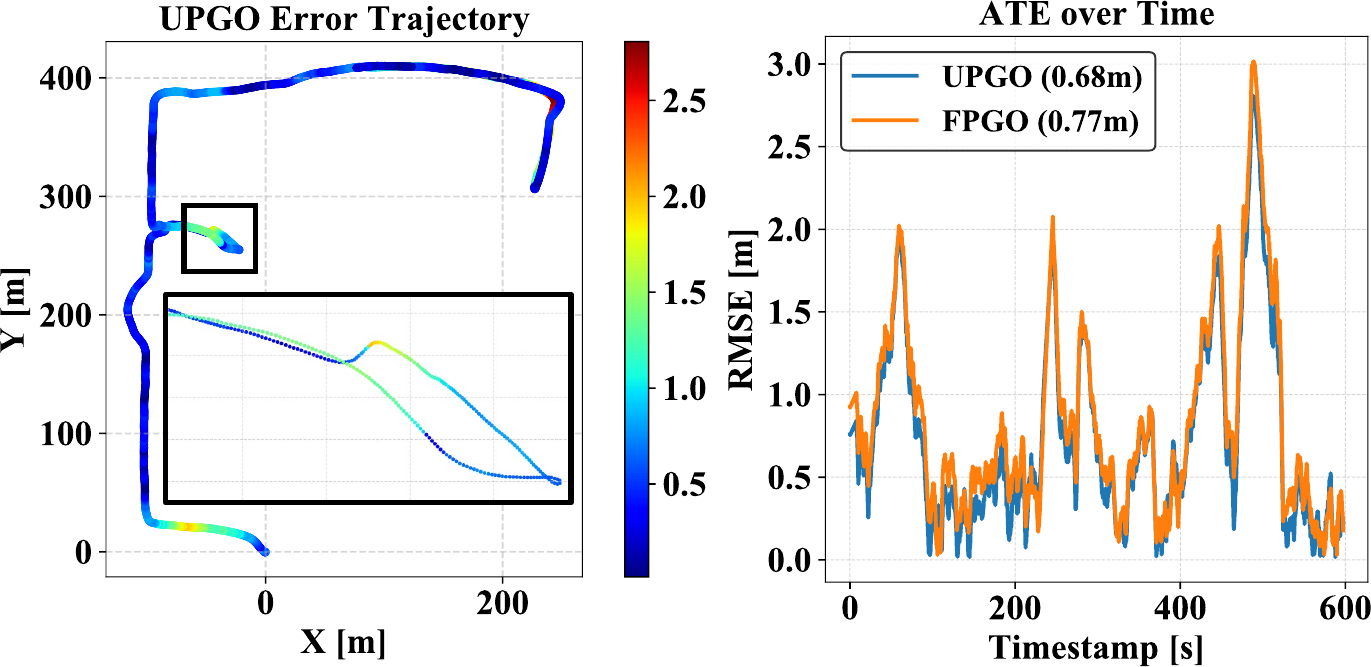}
    \caption{
    \footnotesize
    Time-varying ATE of UPGO and FPGO on the \texttt{CP3} sequence. \textbf{Left}: UPGO error trajectory with color gradient from yellow (high) to blue (low). \textbf{Right}: Temporal ATE variation (XY plane). }
    \label{fig:uncertaity_pk01}
        \vspace{-1.5em}
\end{figure}

\begin{figure*}
    \centering
    \includegraphics[width=0.9\textwidth]{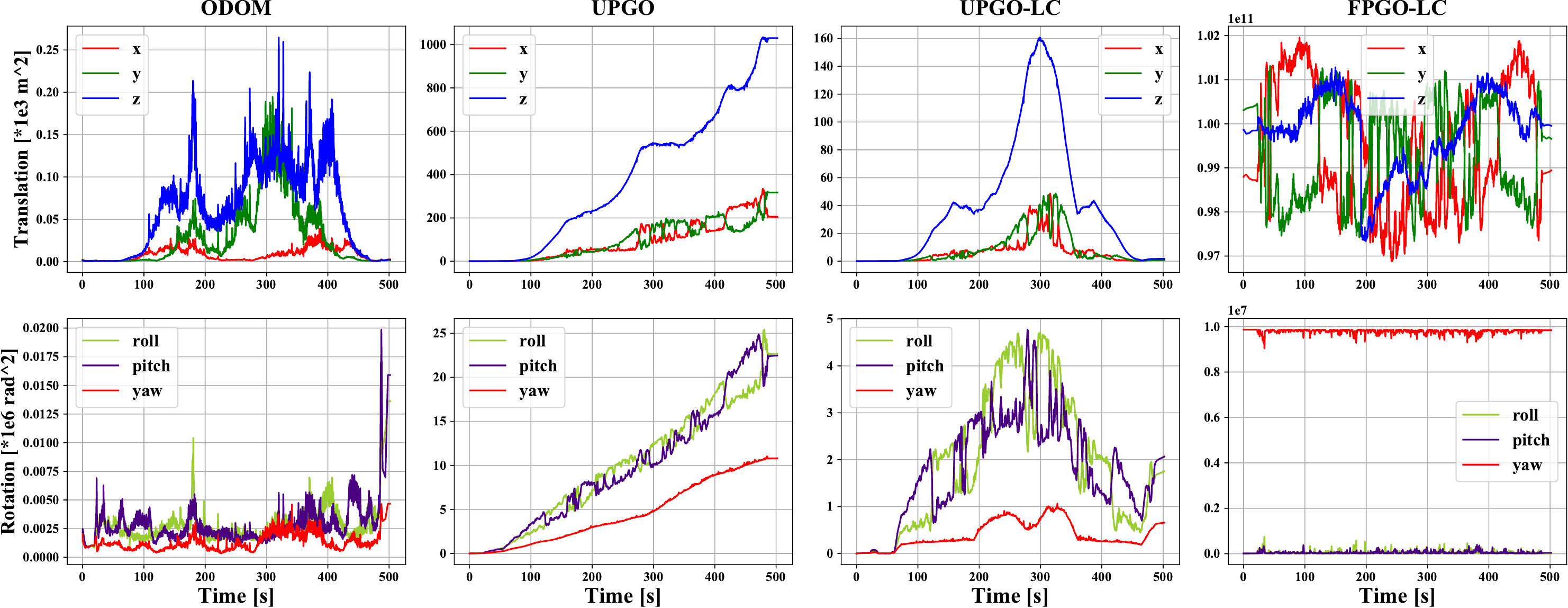}
    \caption{\footnotesize
   Temporal variation of position and orientation variance in the \texttt{PK1}. Columns represent: (1) pure odometry (ODOM), (2) graph optimization without loop closure (UPGO), (3) graph optimization with loop closure (UPGO-LC), and (4) FPGO with loop closure (FPGO-LC). 
   }
    \label{fig:covariance_plots}
           \vspace{-1.5em}
\end{figure*}

\begin{figure}
    \centering
    \includegraphics[width=0.48\textwidth]{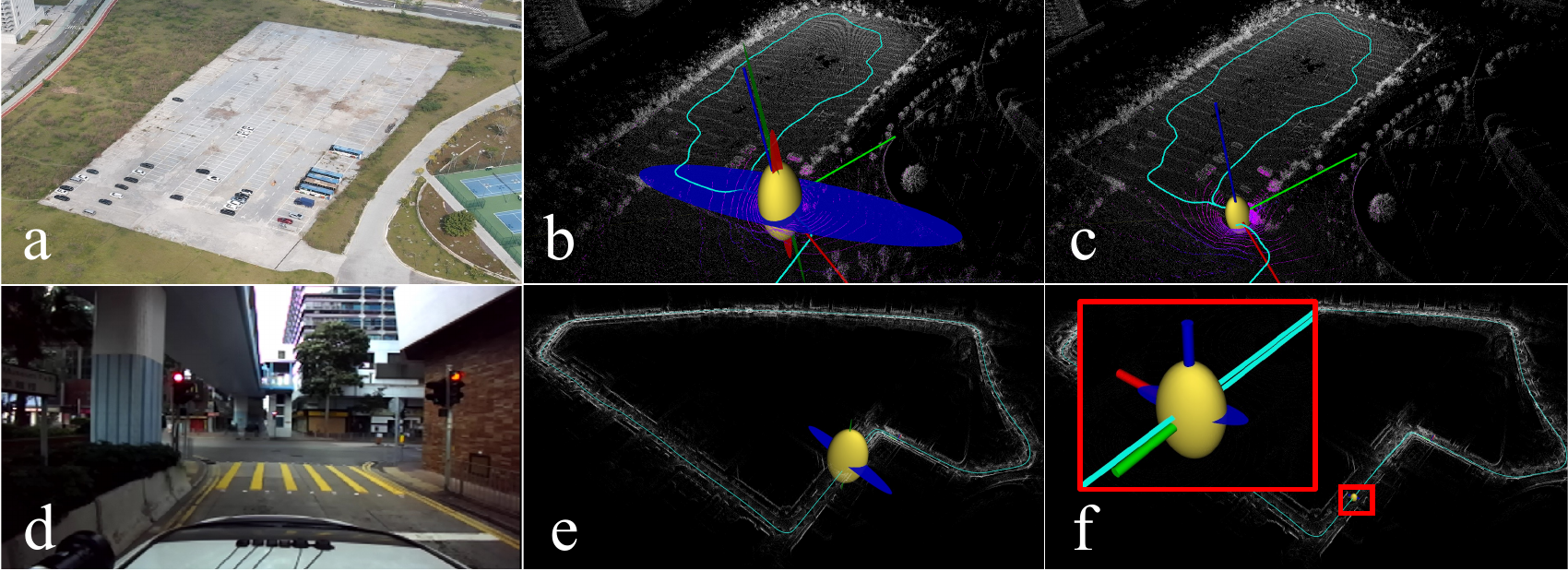}
    \caption{ \footnotesize
    Visualization of pose uncertainty in LiDAR degradation and large-scale urban scenarios (world frame). (a) and (d) depict real scenes from the \texttt{PK1} and \texttt{HST} sequences. In degraded scenarios, (b) and (c) illustrate a significant reduction in estimated pose covariance by UPGO before and after loop closure. Similarly, in large-scale urban scenarios, (e) and (f) show a similar trend in pose variance changes. Yellow ellipsoids represent translational uncertainty, with larger Z-axis uncertainty attributed to lower LiDAR resolution in the Z direction compared to the XY plane. Blue, red, and green ellipses denote rotational uncertainty for the X, Y, and Z axes.
}
    \label{fig:uncertaity}
          \vspace{-1em}
\end{figure}

\subsection{Uncertainty SLAM Expriments}

\subsubsection{Experiment Design}\label{subsec:experimental_design}

\begin{figure}
    \centering
    \includegraphics[width=0.45\textwidth]{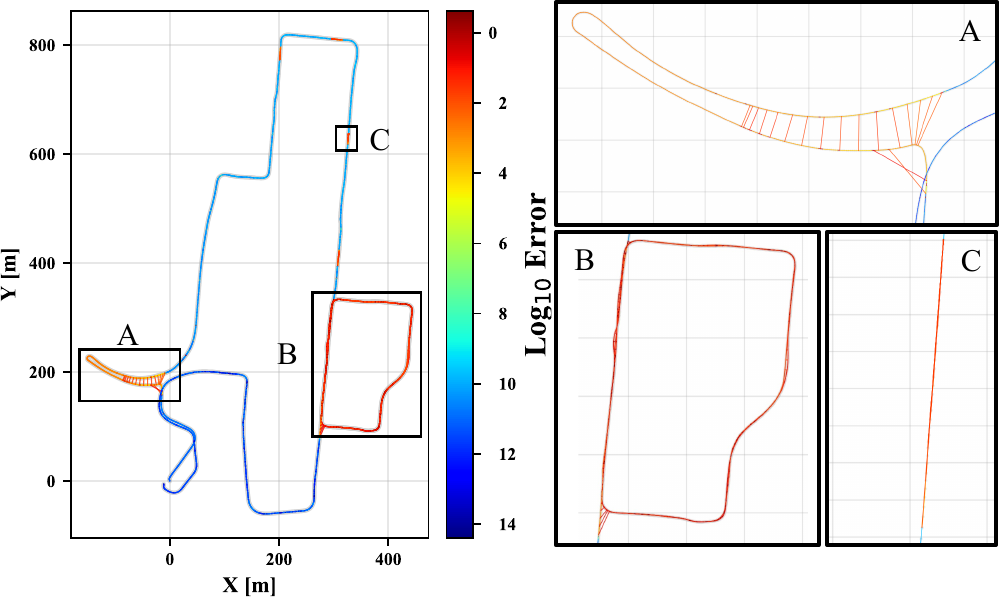}
    \caption{
    \footnotesize
    Error adjustment visualization in UPGO on the \texttt{Whampoa} sequence. Regions A, B, and C represent loop closures at long, medium, and short time duration. UPGO achieves reasonable error adjustment by applying covariance to corresponding edges.}
    \label{fig:graph_error_urban}
          \vspace{-1.5em}
\end{figure}

\begin{figure}
    \centering
    \includegraphics[width=0.45\textwidth]{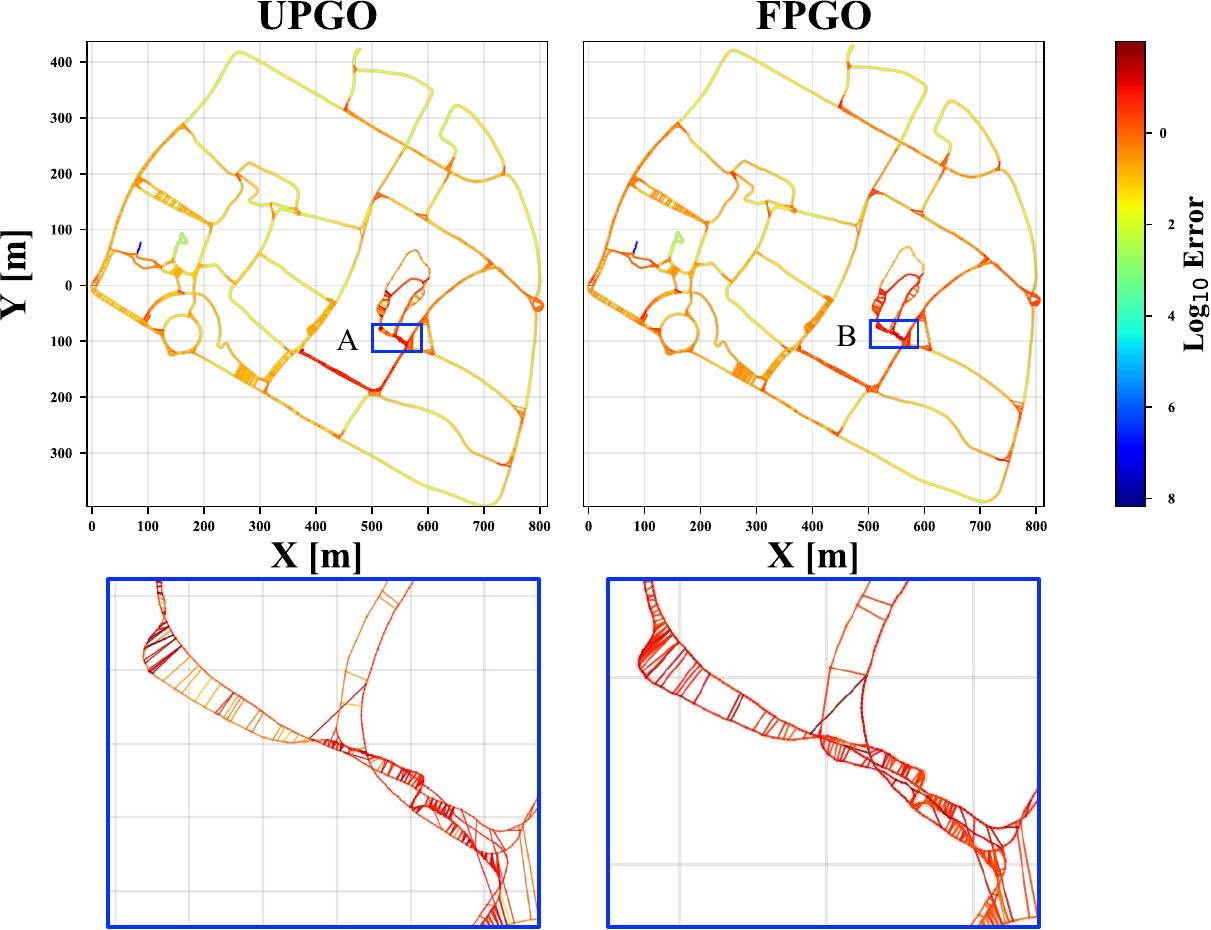}
    \caption{
    \footnotesize
    Error adjustment for incremental mapping of \texttt{PK1} (\SI{0.97}{km}) on \texttt{CP0} (\SI{10.8}{km}). Regions A and B show multiple traversals, with color gradients from red (high) to blue (low) indicating error adjustment. In Region A, loop closure edges from different periods exhibit significant color variations, while Region B shows similar colors, demonstrating the effective of UPGO automatic adjustment through covariance. ATE for UPGO and FPGO are \SI{0.98}{m} and \SI{1.90}{m}, indicating  accuracy improvement in large-scale scenarios.}
    \label{fig:graph_compare}
          \vspace{-1.5em}
\end{figure}

We conduct a series of experiments to validate the performance of the uncertainty model in LiDAR SLAM:
\begin{itemize}
    \item We compare the localization accuracy of FPGO and UPGO in both small and large-scale scenarios. This comparison quantifies the accuracy improvements of the uncertainty model under various environmental conditions (Table~\ref{tab:traj_ape_rpe_half} and Fig.~\ref{fig:uncertaity_pk01}).
    \item We analyze the changes in pose variance by UPGO and FPGO, providing deeper insights into covariance in maintaining the consistency and reliability of the SLAM system (Fig.~\ref{fig:covariance_plots}).
    \item We visualize the error adjustment process of different edges during graph optimization to analyze the impact of different scale loop constraints in UPGO. This method is further applied to large-scale multi-session mapping to demonstrate the improvements and accuracy enhancements brought by UPGO (Fig.~\ref{fig:uncertaity}, Fig.~\ref{fig:graph_error_urban}, and Fig.~\ref{fig:graph_compare}).
\end{itemize}
For UPGO, we set the noise scale to 0.01 for indoor environments and 100 for outdoor environments.

\subsubsection{SLAM Performance}\label{subsec:experimental_results}

We first conducted single-session SLAM experiments across different scales, duration, and diverse scenarios to compare the localization accuracy (ATE) of UPGO and FPGO. To guarantee fairness, all system parameters were kept consistent, except for the uncertainty settings. Our proposed UPGO achieved better results in nearly all tested sequences, although the improvements in some scenarios were minimal, as detailed in Table~\ref{tab:traj_ape_rpe_half}.

In larger scenarios, such as the MS-dataset and Urban-Nav dataset, UPGO significantly improved ATE. However, FPGO faced ill-posed situations leading to graph optimization failures when set with unrealistic noise parameters, as seen in sequences like \texttt{CC1} and \texttt{Mongkok}. This highlights the robustness  and high accuracy of UPGO with the adjustment of a single parameter. Fig.~\ref{fig:ill_pose} further demonstrates how inappropriate noise parameter settings in FPGO result in biased pose trajectories in loop closure regions, indicating local optima in graph optimization and eventual mapping failure. Conversely, UPGO successfully completes the optimization (subfigures (c) and (d)).
We also conducted experiments in smaller scenarios, such as the Newer College and FusionPortable datasets, where ATE improvements were limited. This is because UPGO primarily relies on loop closure constraints to eliminate cumulative errors, while in smaller scenarios, the map fully covers the motion area, creating an "implicit loop closure" that limits the accuracy improvement from loop closures and may introduce larger errors. However, complex scenarios involving abrupt movements (\texttt{quad-hard}) and degraded corridors (\texttt{long\_corridor}) still show significant accuracy improvements.
Moreover, UPGO demonstrated better accuracy across various sensor combinations in different datasets, showcasing its robustness. We compared the accuracy changes brought by UPGO and FPGO throughout the SLAM process on the \texttt{CP3} dataset. Fig.~\ref{fig:uncertaity_pk01} presents the detailed results. In the left subfigure, the small loop closure area shows a noticeable decrease in ATE, while the right image depicts the error variation of UPGO throughout the process, consistently outperforming FPGO.
These experiments demonstrate the superior accuracy and robustness of UPGO, especially in large-scale, complex motion scenarios.

\subsubsection{Uncertainty Analysis}


We conducted experiments in a degraded open parking lot (\texttt{PK1}), which features loop closure constraints before and after entering the parking area. This scenario clearly illustrates the changes in pose covariance, as detailed in Fig.~\ref{fig:covariance_plots}.
We compared four cases: pure odometry (ODOM), UPGO without loop closures, UPGO with loop closures (UPGO-LC), and FPGO with loop closures (FPGO-LC). The comparison focused on the temporal variations in translation and rotation variances. Notably, we described uncertainty in the world frame.
ODOM, using FAST-LIO2 \cite{xu2022fast}, considers only single pose and covariance estimate, primarily reflecting local point cloud registration results based on the LiDAR measurement model (Section~\ref{subsec:iekf}). This explains the large variance range observed in the first column of the two plots.
In the second column, both translation and rotation variances increase monotonically over time, reflecting the cumulative error in odometry in the back-end.
In the third column, introducing loop closures, UPGO-LC shows a sharp decrease in translation and rotation variances, eventually converging to very low values. This validates the direct impact of loop closures on reducing covariance.
In the fourth column, the rotation covariance scale differences in FPGO are significant. FPGO-LC does not exhibit any discernible trend in translation variance, and its yaw variance is significantly larger than roll and pitch, indicating limited practical significance.
These observations underscore the practical applicability of UPGO in real-world scenarios.
Fig.~\ref{fig:uncertaity} further visualizes the covariance ellipses of pose uncertainties before and after loop closures in two different scenarios. In Fig.~\ref{fig:uncertaity}(a) and Fig.~\ref{fig:uncertaity}(b), depicting movements in an open parking lot (Fig.~\ref{fig:covariance_plots}), both translational and rotational covariance significantly decrease after the loop closures. In (c) and (d), depicting a \SI{4.5}{km} urban scenario in \texttt{Whampoa}, a similar trend is observed. These results indicate that the covariance changes in UPGO match our expectations.

In large-scale environments, loop closure is the most effective means to eliminate accumulated errors. However, due to relative constraint errors and improper noise settings, FPGO often suffers from ill-posed graph optimization. In small-scale environments, local maps in odometry may cover the entire area, creating an "implicit loop closure" that limits the benefits of explicit loop closure. Consequently, UPGO offers marginal improvements over FPGO in single-session mapping of such small environments and can sometimes result in negative optimization, as evidenced by experiments on the New College dataset in Table~\ref{tab:traj_ape_rpe_half}. However, in multi-session mapping, UPGO significantly improves performance by effectively handling uncertainty, as demonstrated by the long-duration multi-session results in Table~\ref{tab:overall_performance_comparison}.

\begin{figure}
    \centering
    \includegraphics[width=0.4\textwidth]{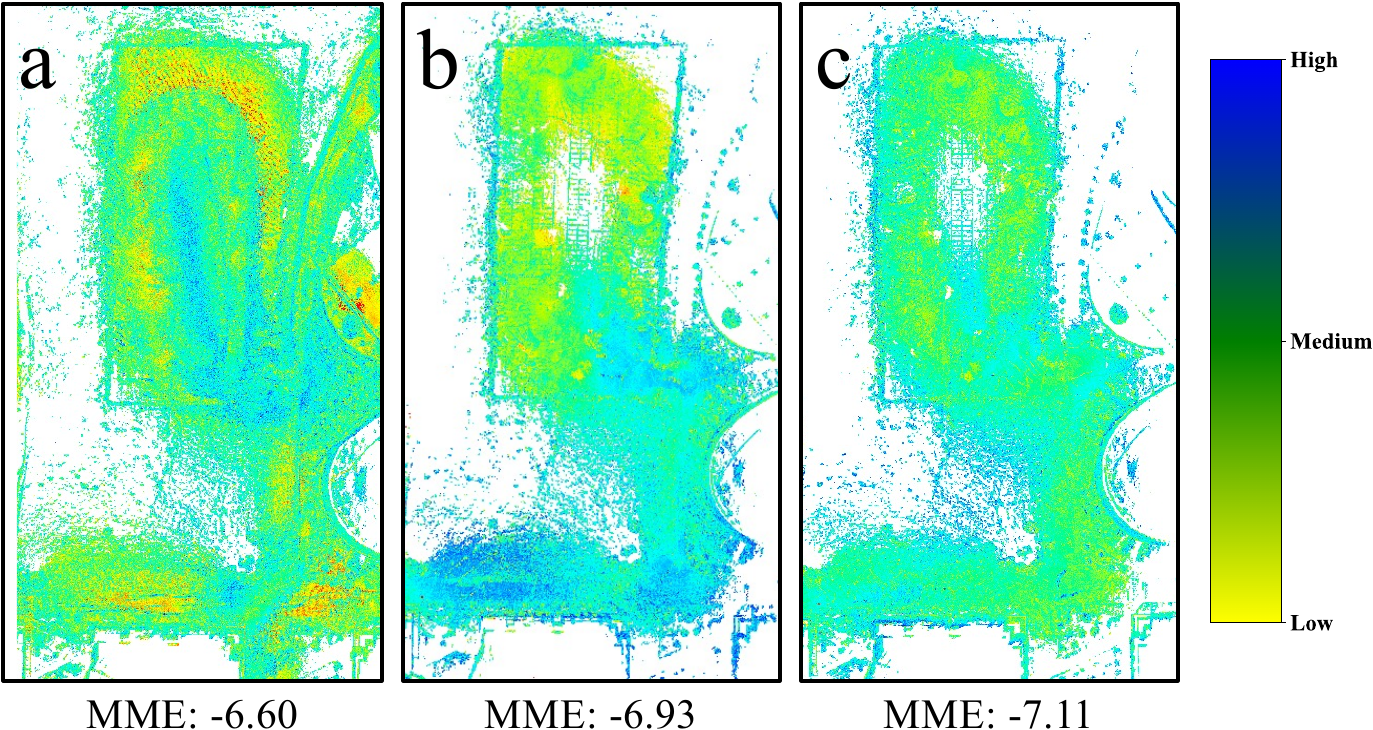  }
\caption{
\footnotesize
Map MME comparison on the \texttt{CP0-PK1} sequences. (a) M2M. (b) F2F. (c) MS. Darker shades of blue indicate higher map consistency.}
    \label{fig:mme_comparison}
          \vspace{-1.5em}
\end{figure}
To verify the advantages of our uncertainty handling method in large-scale environments, we conducted single-session experiments on the Urban-Nav dataset, demonstrating error adjustment performance from graph optimization at different loop closure scales. We then extended the experiments to a \SI{10}{km}-scale multi-session mapping scenario, analyzing the error distribution results after graph optimization. 
Fig.~\ref{fig:graph_error_urban} shows the error adjustment of pose graph edges of UPGO for the \texttt{Whampoa} sequence. The color bar from red to blue represents the magnitude of error assigned to each edge, with deeper red indicating larger adjustments post-optimization. Regions A, B, and C illustrate large, medium, and small loop closure areas. In large-scale graph optimization, loop closures distribute errors through the information matrices, resulting in significant adjustments to node poses within the loop closure region. In Region A, the loop closure spans almost the entire trajectory (A$\rightarrow$B), leading to a noticeable color change to light blue between A and B. In Region B, a medium-sized loop closure adjusts poses within its area, showing a distinct color difference from Region A. Region C shows a stationary loop closure with larger error adjustments post-optimization.
Fig.~\ref{fig:graph_compare} further illustrates the error adjustment of pose graph edges before and after incremental mapping of the \SI{10.8}{km} \texttt{CP0} dataset using \texttt{PK1}. With FPGO, errors are evenly distributed, indicated by minimal color differences among loop closure edges in Region B. In contrast, UPGO shows significant color differences among loop closure edges, highlighting the impact of incorporating uncertainty, resulting in approximately 50\% improvement in ATE.

\begin{table}[t]
\centering
\caption{Comparison of Large-scale Mapping Performance}
\label{tab:overall_performance_comparison}
\renewcommand{\arraystretch}{1.0}
\begin{threeparttable}
\begin{tabular}{@{}lcccccc@{}}
\toprule[0.03cm]
\multirow{2}{*}{\textbf{Comb.}} & \multirow{2}{*}{\textbf{Method}} & \multirow{2}{*}{\textbf{KFR [\%]}} & \multicolumn{4}{c}{\textbf{Metrics}} \\
\cmidrule(lr){4-7}
& & & \textbf{ATE $\downarrow$} & \textbf{AC $\downarrow$} & \textbf{CD $\downarrow$} & \textbf{MME $\downarrow$} \\
\midrule[0.03cm]

\texttt{CP0} & \multicolumn{1}{c}{-} & \multicolumn{1}{c}{-} & \multicolumn{1}{c}{-} & 39.52 & 79.25 & -6.87 \\
\midrule

\multirow{7}{*}{\texttt{P1}} & \multirow{1}{*}{M2M} & - & - & 22.46 & 44.06 & -6.67 \\
\cmidrule(l){2-7}
& \multirow{3}{*}{F2F} & 0-40 & 165.41 & 22.70 & 45.57 & \underline{-7.31} \\
& & 40-70 & 169.32 & 21.59 & 43.79 & -7.26 \\
& & 70-100 & 162.32 & 19.61 & 41.12 & -7.23 \\
\cmidrule(l){2-7}
& \multirow{3}{*}{MS} & 0-40 & \textbf{107.12} & \textbf{15.88} & \textbf{38.49} & \textbf{-7.34} \\
& & 40-70 & \underline{123.95} & 18.12 & 39.96 & -7.27 \\
& & 70-100 & 138.12 & \underline{17.24} & \underline{38.57} & -7.23 \\
\midrule

\multirow{7}{*}{\texttt{P2}} & \multirow{1}{*}{M2M} & - & - & 22.12 & 43.66 & -6.61 \\
\cmidrule(l){2-7}
& \multirow{3}{*}{F2F} & 0-40 & - & 17.65 & 38.80 & -6.94 \\
& & 40-70 & - & 17.00 & 37.14 & -6.93 \\
& & 70-100 & - & 17.28 & 37.72 & -6.92 \\
\cmidrule(l){2-7}
& \multirow{3}{*}{MS} & 0-40 & - & \textbf{12.00} & \textbf{33.01} & \textbf{-7.08} \\
& & 40-70 & - & \underline{13.63} & \underline{34.58} & \underline{-7.04} \\
& & 70-100 & - & 16.49 & 36.88 & -6.91 \\
\midrule

\multirow{7}{*}{\texttt{B2}} & \multirow{1}{*}{M2M} & - & - & 21.39 & 37.83 & -6.59 \\
\cmidrule(l){2-7}
& \multirow{3}{*}{F2F} & 0-40 & \underline{57.95} & 21.01 & 42.43 & -6.92 \\
& & 40-70 & 60.94 & 21.50 & 42.22 & -6.92 \\
& & 70-100 & 66.29 & 21.41 & 41.92 & \underline{-6.94} \\
\cmidrule(l){2-7}
& \multirow{3}{*}{MS} & 0-40 & \textbf{53.96} & \textbf{13.76} & \textbf{33.25} & \textbf{-6.95} \\
& & 40-70 & 59.97 & \underline{17.49} & \underline{37.25} & -6.91 \\
& & 70-100 & 59.78 & 19.59 & 39.88 & -6.91 \\
\midrule

\multirow{6}{*}{\texttt{K1}} & \multirow{1}{*}{M2M} & - & - & 21.70 & 42.94 & -6.60 \\
\cmidrule(l){2-7}
& \multirow{3}{*}{F2F} & 0-40 & \underline{61.33} & 16.88 & 35.75 & -6.91 \\
& & 40-70 & \textbf{57.52} & 15.46 & 33.28 & -6.90 \\
& & 70-100 & 85.01 & 18.12 & 36.20 & -6.93 \\
\cmidrule(l){2-7}
& \multirow{3}{*}{MS} & 0-40 & 76.12 & \textbf{11.15} & \textbf{31.46} & \textbf{-7.11} \\
& & 40-70 & 77.12 & \underline{15.29} & \underline{33.47} & \underline{-7.08} \\
& & 70-100 & 85.01 & 17.09 & 36.55 & -6.86 \\
\bottomrule[0.03cm]
\end{tabular}
\begin{tablenotes}[flushleft]
\footnotesize
 \item ATE, AC, and CD are measured in \SI{}{cm}. '-' indicates unavailable results. \textbf{Bold} show the best results, while \underline{underlined} denote the second-best.
    \item \texttt{P1}: \texttt{CP0}-\texttt{CP1} sequences; \texttt{P2}: \texttt{CP0}-\texttt{CP2} sequences; \texttt{B2}: \texttt{CP0}-\texttt{RB2} sequences; \texttt{K1}: \texttt{CP0}-\texttt{PK1} sequences.
\end{tablenotes}
\end{threeparttable}
\vspace{-0.5em}
\end{table}

\begin{figure}
    \centering
    \includegraphics[width=0.4\textwidth]{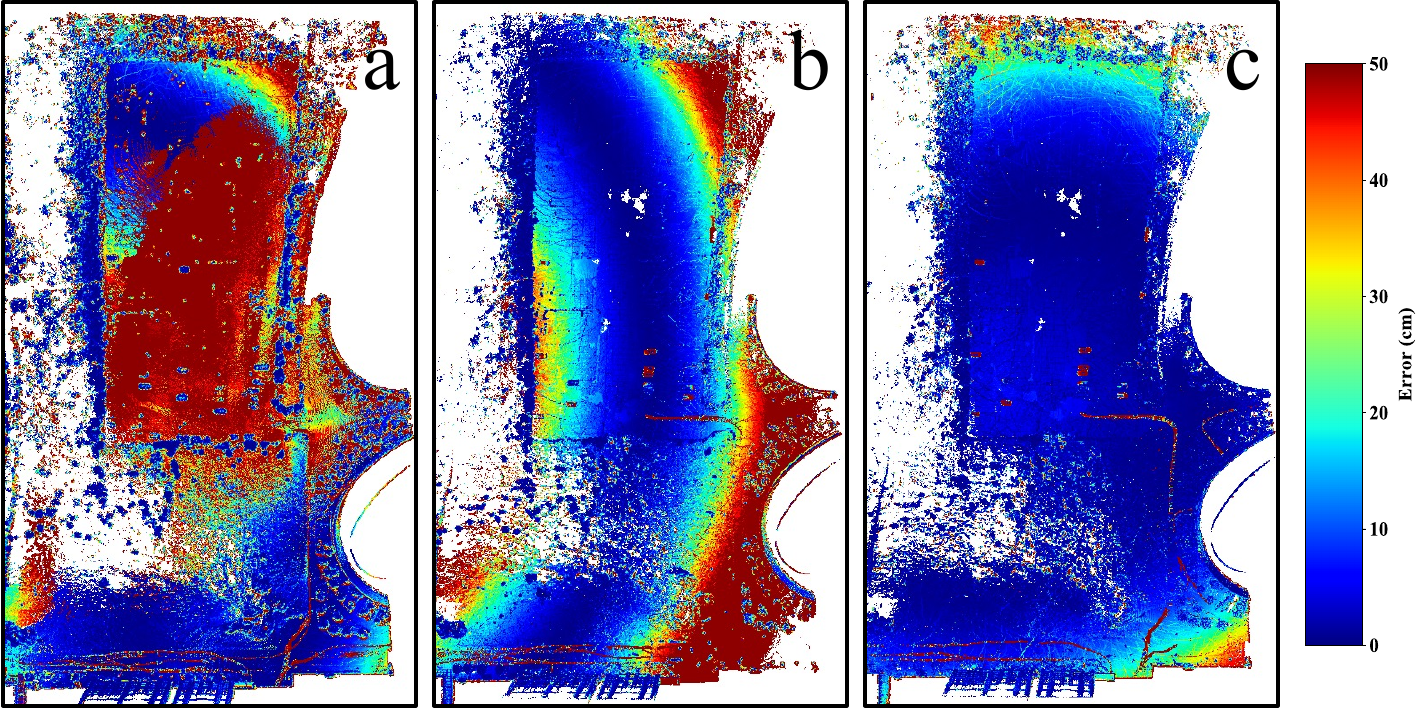}
    \caption{
    \footnotesize
    Map accuracy comparison on \texttt{CP0-PK1} sequences: (a) M2M, (b) F2F, (c) MS. Color scale: red (high error) to blue (low error).}
    \label{fig:error_map_pk01}
          \vspace{-2em}
\end{figure}

\section{Systematic Experimental Results}\label{sec:sec_sys_exp}

\subsection{System Experiments}
\subsubsection{Experiment Design}
\begin{figure*}
    \centering
    \includegraphics[width=0.9\textwidth]{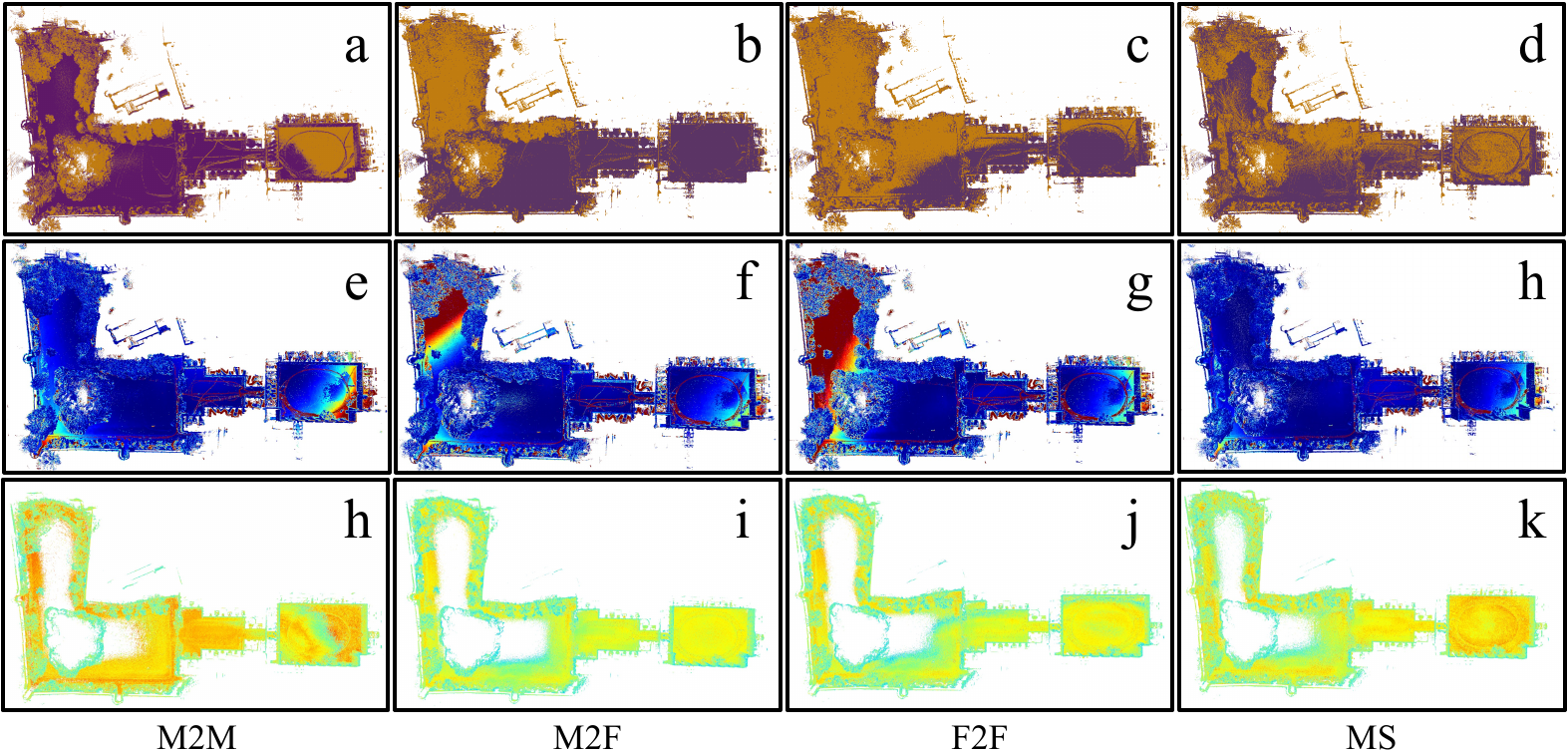}
    \caption{ 
    \footnotesize
     Visualization of map evaluation results for MS-Mapping and baseline algorithms using the \texttt{short-exp11} sequence (\texttt{S2}) on the \texttt{parkland0} (\texttt{S1}), as shown in Table~\ref{tab:algorithm-comparison-nc}. \textbf{(a)-(d)} illustrate the map outputs of various algorithms, with purple indicating point clouds from \texttt{parkland0} and yellow indicating point clouds from \texttt{short-exp11}. \textbf{(e)-(f)} depict the map AC visualization, using a color bar consistent with Fig.~\ref{fig:error_map_pk01}, where the maximum error range is \SI{0.5}{m}(\rt{red areas}). \textbf{(h)-(k)} show the map MME visualization for different algorithms, the colorbar is the same as Fig.~\ref{fig:mme_comparison}.}
    \label{fig:ms_bs_conpare_nc}
    \vspace{-1.5em}
\end{figure*}

\begin{figure*}
    \centering
    \includegraphics[width=0.8\textwidth]{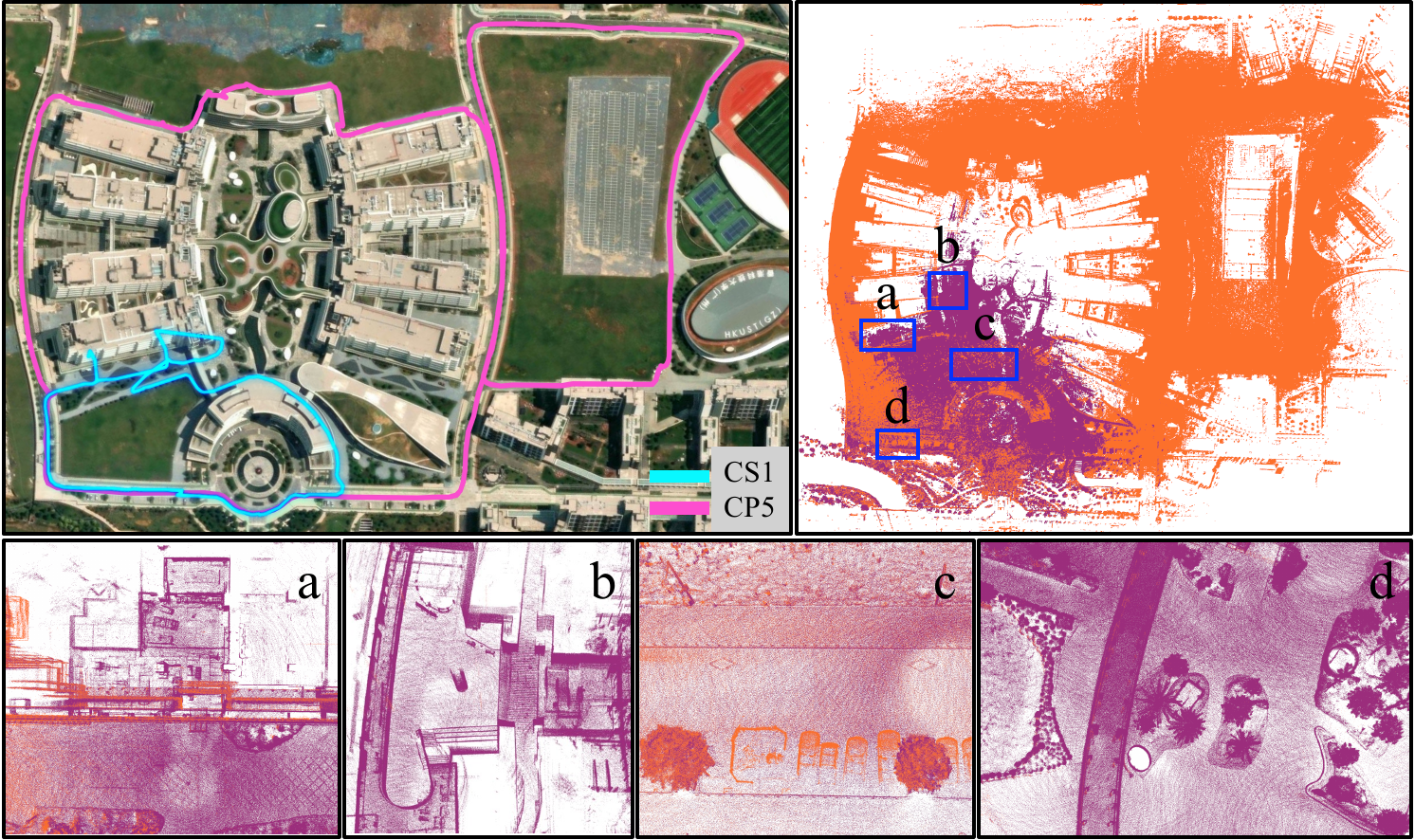}
\caption{ 
\footnotesize
Incremental mapping performance on \texttt{CP5-CS1} (\SI{4.3}{km}). (a) Transition from outdoor to indoor. (b) Transition from indoor to outdoor. (c) Non-overlapping region of \texttt{CS1}. (d) Overlapping region of \texttt{CP5-CS1}.}
    \label{fig:transitions_exp}
          \vspace{-1.5em}
\end{figure*}

\begin{figure}
    \centering
    \includegraphics[width=0.45\textwidth]{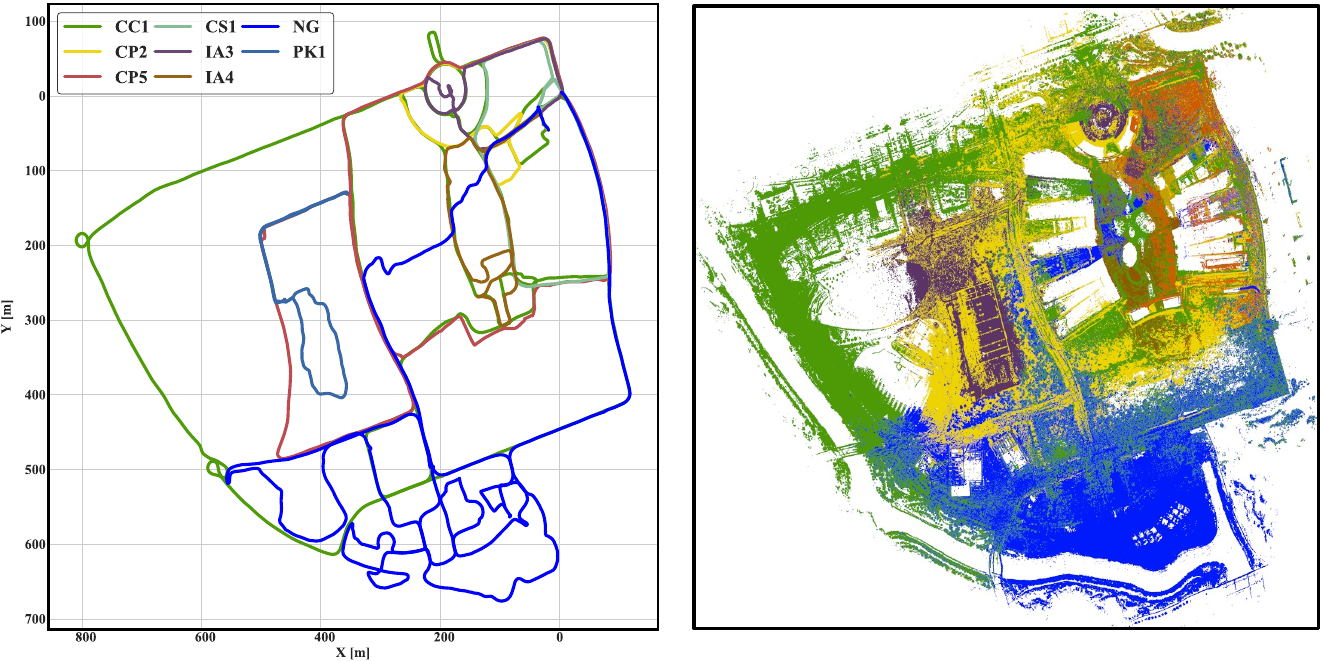}
    \caption{
    \footnotesize
    Map merging results for 8 session data from the MS-Dataset. \textbf{Left}: the merged trajectory using MS-Mapping algorithm. \textbf{Right}: The merged map, different colors represent different sessions.}
    \label{fig:8session}
          \vspace{-2em}
\end{figure}

To thoroughly evaluate the MS-Mapping system, we conducted several distinct experiments:
\begin{itemize}
    \item Incremental mapping on the large-scale \texttt{CP0} dataset using four additional data sequences, assessing changes in localization and mapping under different KFR to validate the performance of MS-Mapping (Table~\ref{tab:overall_performance_comparison}, Fig.~\ref{fig:error_map_pk01}, Fig.~\ref{fig:mme_comparison}).
    \item  Cross-incremental mapping experiments on the \texttt{RB2} and \texttt{PK1} data sequences, comparing the accuracy of the Base Map (BM), Isolated Map (IM), and Composite Map (CM) at different KFR to evaluate the consistency and robustness of the MS-Mapping system (Table~\ref{tab:cross_session_performance_comparison}).
    \item Incremental mapping on the \texttt{parkland0} and \texttt{short-exp11} datasets, focusing on mapping in small-scale scenes with long duration and high overlap areas. We comprehensively analyzed the differences in map performance between the MS-Mapping and baseline algorithms for single-session and multi-session mapping, considering both individual sequence maps and the overall map (Table~\ref{tab:algorithm-comparison-nc} and Fig.~\ref{fig:ms_bs_conpare_nc}).
    \item Incremental mapping experiments in scenarios with large-scale indoor and outdoor transitions, using two different platforms with the same set of parameters, to demonstrate the broad applicability of the MS algorithm (Fig.~\ref{fig:transitions_exp}).
   \item  We conducted large-scale mapping experiments with data from eight sessions to demonstrate the incremental mapping capabilities of the MS-Mapping algorithm (Fig.~\ref{fig:8session}).
\end{itemize}
For clarity, when we perform incremental mapping on the \texttt{S1} sequence using new data from the \texttt{S2} sequence, we refer to this data as "\texttt{S1}-\texttt{S2}".

\begin{table}[t]
\centering
\caption{Comparison of Cross Multi-Session Mapping.}
\label{tab:cross_session_performance_comparison}
\renewcommand{\arraystretch}{1.0}
\begin{threeparttable}
\begin{tabular}{@{}llcccccc@{}}
\toprule[0.03cm]
\textbf{Comb.} & \textbf{Type} & \textbf{KFR [\%]} & \textbf{ATE $\downarrow$} & \textbf{AC $\downarrow$} & \textbf{CD $\downarrow$} & \textbf{MME $\downarrow$} \\
\midrule[0.03cm]
\multirow{6}{*}{\texttt{R1}} & \multirow{1}{*}{BM} & 55.36 & \cellcolor{Gray}59.16 & \cellcolor{Gray}19.27 & \cellcolor{Gray}39.36 & \cellcolor{Gray}-6.90 \\
\cmidrule(l){2-7}
& \multirow{3}{*}{IM} & 0-40 & \textbf{75.10} & \textbf{10.60} & \textbf{30.52} & \textbf{-6.90} \\
& & 40-70 & 90.86 & \underline{16.24} & \underline{34.80} & \underline{-6.77} \\
& & 70-100 & \underline{82.69} & \cellcolor{Gray}17.59 & \cellcolor{Gray}35.14 & \cellcolor{Gray}-6.75 \\
\cmidrule(l){2-7}
& \multirow{3}{*}{CM} & 0-40 & \textbf{75.10} & 19.70 & 40.52 & -6.75 \\
& & 40-70 & 90.86 & 19.11 & 39.39 & -6.68 \\
& & 70-100 & 82.70 & \cellcolor{Gray}20.81 & \cellcolor{Gray}41.54 & \cellcolor{Gray}-6.71 \\
\midrule

\multirow{6}{*}{\texttt{K2}} & \multirow{1}{*}{BM} & 84.09 & \cellcolor{Gray}59.51 & \cellcolor{Gray}17.56 & \cellcolor{Gray}36.20 & \cellcolor{Gray}-5.75 \\
\cmidrule(l){2-7}
& \multirow{3}{*}{IM} & 0-40 & \textbf{52.10} & \textbf{13.59} & \underline{35.03} & -6.79 \\
& & 40-70 & 65.89 & 16.13 & 35.68 & \underline{-6.90} \\
& & 70-100 & 77.71 & 15.39 & 34.65 & \textbf{-6.93} \\
\cmidrule(l){2-7}
& \multirow{3}{*}{CM} & 0-40 & \underline{52.11} & \underline{14.60} & 34.33 & -6.67 \\
& & 40-70 & 65.89 & 17.32 & 36.29 & -6.76 \\
& & 70-100 & 77.71 & 18.27 & 37.63 & -6.78 \\
\bottomrule[0.03cm]
\end{tabular}
\begin{tablenotes}[flushleft]
\footnotesize
\item ATE, AC and CD are measured in \SI{}{cm}.
\textbf{Bold} shows the best results, while \underline{underlined} denotes the second-best. \texttt{R1}: \texttt{RB2}-\texttt{PK1} sequences; \texttt{K2}: \texttt{PK1}-\texttt{RB2} sequences.
\end{tablenotes}
\end{threeparttable}
\vspace{-1.5em}
\end{table}

\begin{table}[t]
\centering
\caption{Comparison of Experimental Results for Incremental Mapping on \texttt{parkland0} (\texttt{S1}) using \texttt{short-exp11} (\texttt{S2}).}
\renewcommand{\arraystretch}{1.0}
\begin{threeparttable}
\begin{tabular}{@{}ll ccc cc@{}}
\toprule[0.03cm]
\multicolumn{2}{c}{} & \multicolumn{2}{c}{\textbf{Individual}} & \multicolumn{3}{c}{\textbf{Combination}} \\
\cmidrule(lr){3-4} \cmidrule(lr){5-7}
\textbf{Method} & \textbf{Metric} & \texttt{S1} & \texttt{S2} & \texttt{S1} & \texttt{S2} & \texttt{S1+S2} \\
\midrule[0.03cm]
\multirow{4}{*}{M2M} & \textbf{ATE [\si{cm}] $\downarrow$} & 28.24 & 46.15 &-- & -- & -- \\
 & \textbf{AC [\si{cm}] $\downarrow$} & 15.42 & 16.10 & 15.42 & \underline{16.10} & 16.47 \\
  & \textbf{CD [\si{cm}] $\downarrow$} & 31.46 & 33.17 & 31.46 & 33.17 & 30.91 \\
 & \textbf{MME} $\downarrow$ & -8.74 &-8.51 & \textbf{-8.75} & \textbf{-8.51} & -8.49 \\
\midrule
\multirow{3}{*}{M2F} & \textbf{ATE [\si{cm}] $\downarrow$} & 28.24 & 46.15 & \textbf{26.98} & \underline{48.92} & -- \\
 & \textbf{AC [\si{cm}] $\downarrow$} & 15.42 & 16.10 & \textbf{14.40} & 17.21 & 16.49 \\
 & \textbf{CD [\si{cm}] $\downarrow$} & 31.46 & 33.17 & \textbf{28.81} & 35.12 & 29.47 \\
 & \textbf{MME} $\downarrow$ & -8.74 & -8.51 & \underline{-8.74} & -8.38 & \textbf{-8.63} \\
\midrule
\multirow{3}{*}{F2F} & \textbf{ATE [\si{cm}] $\downarrow$} & 28.24 & 46.15 & \textbf{26.98} & 49.90 & -- \\
 & \textbf{AC [\si{cm}] $\downarrow$} & 15.42 & 16.10 & \underline{14.41} & 16.50 & \underline{16.02} \\
 & \textbf{CD [\si{cm}] $\downarrow$} & 31.46 & 33.17 & \textbf{28.81} & \underline{32.89} & \underline{29.23} \\
 & \textbf{MME} $\downarrow$ & -8.74 & -8.51 & -8.73 & -8.39 & \underline{-8.59} \\
\midrule
\multirow{3}{*}{MS} 
& \textbf{ATE [\si{cm}] $\downarrow$} &  \textbf{26.25} & \textbf{45.05} & \underline{29.15} & \textbf{39.89} & -- \\
 & \textbf{AC [\si{cm}] $\downarrow$} & \textbf{14.19} & \textbf{15.99} & 15.47 & \textbf{14.90} & \textbf{15.40} \\
 & \textbf{CD [\si{cm}] $\downarrow$} & \textbf{28.34} & \textbf{32.98} & \underline{30.71} & \textbf{31.72} & \textbf{29.06} \\
 & \textbf{MME} $\downarrow$ & -8.59 & -8.46 & \underline{-8.74} & \underline{-8.46} & -8.57 \\
\bottomrule[0.03cm]
\end{tabular}
\begin{tablenotes}[flushleft]
\footnotesize
\item \textbf{Bold} shows the best results, while \underline{underlined} denotes the second-best.
\end{tablenotes}
\end{threeparttable}
\label{tab:algorithm-comparison-nc}
  \vspace{-1.5em}
\end{table}

\subsubsection{Large-scale Mapping Accuracy Evaluation}


To guarantee the scene was sufficiently large and capable of direct map accuracy evaluation, we conducted incremental mapping experiments using the \texttt{CP0} sequence as the base, supplemented by several other indoor and outdoor sequences (\texttt{CP1}, \texttt{CP2}, \texttt{RB2}, \texttt{PK1}) across different KFRs. Due to the difficulty in obtaining ground truth for the \texttt{CP0} sequence trajectory, and given that most new session data sequences have accurate ground truth trajectories, we only evaluated the overall trajectory and map accuracy in the new session regions after incremental mapping. Detailed experimental results are presented in Table~\ref{tab:overall_performance_comparison}.

Among the four sets of experiments, the M2M algorithm demonstrated superior map accuracy compared to F2F and MS algorithms. This is because the base map constructed using the \texttt{CP0} sequence was much larger, while the new session areas for incremental mapping were relatively smaller. Using CloudCompare to register these maps together may have caused some regions of the new session maps to exhibit warping, as shown in Fig.~\ref{fig:error_map_pk01}(a).
In contrast, the MS algorithm outperformed F2F across all KFR ranges in terms of map accuracy and Chamfer Distance, showcasing its accuracy advantage. Additionally, MS achieved the best results in MME, indicating better local map continuity. Fig.~\ref{fig:mme_comparison} visualizes the MME comparison in the parking lot area, where the MS algorithm displayed a balanced blue distribution, demonstrating superior map continuity over the other two algorithms.
However, in some scenarios, such as \texttt{CP0-PK1}, MS did not achieve the best localization accuracy, indicating that high localization accuracy in large-scale mapping does not necessarily guarantee better map accuracy. Comparing MS at increasing KFRs, both localization and map accuracy consistently declined. This is because selecting more keyframes for map merging introduces more errors. Therefore, setting an appropriate keyframe threshold is crucial for mapping accuracy, but fewer keyframes might result in incomplete maps, requiring a balance. The experimental results in Fig.~\ref{fig:keyframe} guide us in setting reasonable keyframe thresholds.

\subsubsection{Cross-Incremental Mapping Evaluation}

To evaluate the consistency and robustness of the proposed MS-Mapping system, we conducted cross-incremental mapping experiments on the \texttt{RB2} and \texttt{PK1} sequences. Table~\ref{tab:cross_session_performance_comparison} presents the accuracy changes of the base map (BM), the new session map (IM), and the merged map (CM) when incrementally mapping \texttt{PK1} on \texttt{RB2}  and vice versa.
IM refers to the newly added regions after incremental mapping, excluding the BM. Therefore, its localization accuracy remains nearly unchanged compared to the corresponding CM localization accuracy. In both datasets, under different KFRs, the IM results consistently outperform CM, revealing a slight decrease in  accuracy of the merged map, aligning with practical situations.
When comparing the IM in the \texttt{RB2}-\texttt{PK1} data with the BM in the \texttt{PK1}-\texttt{RB2}, and the IM in the \texttt{PK1}-\texttt{RB2} with the BM in the \texttt{RB2}-\texttt{PK1}, the IM accuracy is consistently better. This indicates that by excluding the interference of overlapping regions, MS can improve the map accuracy of the target area during incremental mapping by leveraging associations constructed from old session data compared to single-session mapping.

\subsubsection{Small-Scale and Long-Duration Incremental Mapping}

We conducted comprehensive experiments on the small-scale New College dataset, which includes long duration and diverse environments such as buildings and woods, to show the differences between single and multi-session mapping. 
Using consistent parameters, we incrementally updated the map from \texttt{parkland0} (\texttt{S1}) with \texttt{short-exp11} (\texttt{S2}). The \texttt{parkland0} data was collected using Ouster-OS$0$ in 2021, while \texttt{short-exp11} was collected using Ouster-OS$1$ in 2020, with minor seasonal differences in the wooded and grassy areas. Detailed quantitative results are presented in Table~\ref{tab:algorithm-comparison-nc}.
The MS algorithm showed significant improvements in AC and CD metrics, based on calculations using over 10 million points with a point cloud overlap rate exceeding 99\%, ensuring reliability. 
Although MS did not achieve the best result in MME, the difference was minimal. 
When evaluating the regional maps of individual sessions after merging, MS achieved nearly a 20\% improvement in localization accuracy for \texttt{S2}, with similar trends in AC and CD metrics. However, using UPGO slightly affected the accuracy of the \texttt{S1} regional map. M2F achieved the best accuracy in the \texttt{S1} regional map, possibly due to optimization based on the overall \texttt{S1} map prior, resulting in minimal accuracy difference post-merging. F2F, using the prior map frames for loop closure, slightly adjusted the \texttt{S1} region.
For single-session evaluation, MS showed better AC and CD metrics in the UPGO portion compared to baseline algorithms (FPGO). The superior performance of MS, which utilizes different sensor combinations for map merging, highlights its robustness. 
When single-session mapping falls short, MS-Mapping can leverage additional data sequences to enhance overall map accuracy.
Figure~\ref{fig:ms_bs_conpare_nc} illustrates the visual results of the MS and baseline algorithms, as detailed in Table~\ref{tab:algorithm-comparison-nc}. The yellow point clouds in MS (subfigure (d)) are uniformly distributed, whereas the baseline algorithms show regional distribution, indicating better map consistency with MS. M2F and F2F exhibit significant errors in the left woodland area, while M2M shows high errors in the right building area. Only MS maintains high accuracy across the entire map (subfigure (h)).
Figure~\ref{fig:in_map_vi} further demonstrates the map merging results of the MS algorithm, as detailed in Table~\ref{tab:algorithm-comparison-nc}. 
With just a single parameter set, the MS algorithm achieves excellent results in mapping both indoor and outdoor environments during map merging.
\begin{figure}
    \centering
    \includegraphics[width=0.45\textwidth]{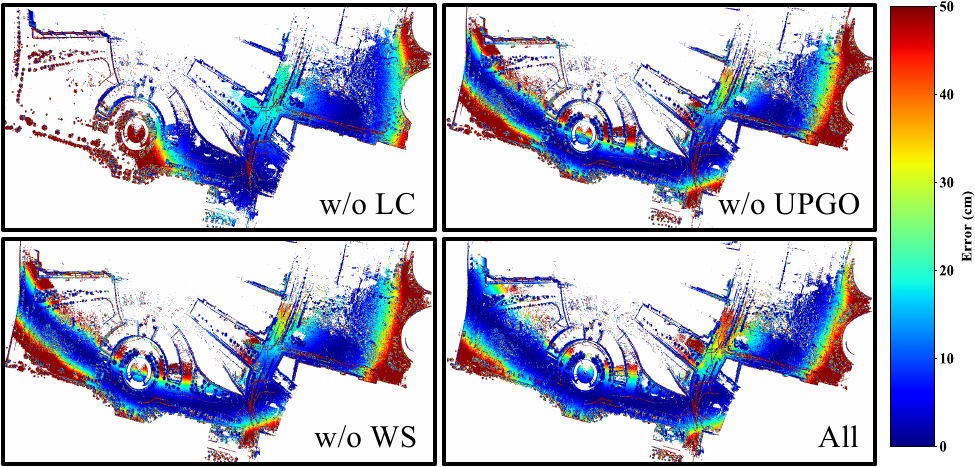}
    \caption{\textbf{
    \footnotesize
    Map error visualization of the ablation study on the \texttt{PK1-RB2} dataset}. (a) W/O LC, (b) W/O UPGO, (c) W/O WS, (d) All. The colorbar ranges from red to green to blue, representing error levels from 0 to \SI{50}{cm}. Bluer regions indicate lower errors. }
    \label{fig:ablation_map}
          \vspace{-2em}
\end{figure}

\begin{table}[!t]
\centering
\caption{Ablation study on the \texttt{RB2}-\texttt{PK1} dataset.}
\begin{threeparttable}
\renewcommand\arraystretch{1.0}
\begin{tabular}{lcccc}
\toprule[0.03cm]
\textbf{Method} & \textbf{ATE [\SI{}{cm}]} $\downarrow$ & \textbf{AC [\SI{}{cm}]} $\downarrow$ & \textbf{CD [\SI{}{cm}]} $\downarrow$ & \textbf{MME} $\downarrow$ \\
\midrule[0.03cm]
w/o LC & 62.4   & 21.3 & 42.1 & \textbf{-5.90} \\
w/o WS & \underline{52.5}   & 21.0 & 41.5& -5.65\\
w/o UPGO & 57.2   & \underline{19.5} & \underline{39.3}& -5.67 \\
All   & \textbf{48.3}  & \textbf{19.1} & \textbf{38.9}& \underline{-5.74} \\
\bottomrule[0.03cm]
\end{tabular}
\begin{tablenotes}[flushleft]
\item \textbf{Bold} shows the best results, while \underline{underlined} denotes the second-best.
\end{tablenotes}
\end{threeparttable}
\label{tab:ablation_experiment}
    \vspace{-1.5em}
\end{table}

\subsubsection{Multi-Environment and Multi-Sensor Incremental Mapping}

Fig.~\ref{fig:transitions_exp} depicts a typical scenario where MS-Mapping is used for incremental mapping across different environments and sensor combinations. Based on \texttt{CP5}, we incrementally mapped with \texttt{CS1} over a scene spanning more than \SI{4.3}{km}, including transitions from large-scale outdoor areas to indoor environments and back to outdoor settings. Subfigure (a), (b), (c), and (d) illustrate the map details from outdoor to indoor, indoor to outdoor, overlapping regions, and non-overlapping regions, with the orange and purple point clouds representing the \texttt{CP5} and \texttt{CS1} maps. This demonstrates the capability of our MS-Mapping to expand the map across diverse environments and sensor combinations.
To further demonstrate the large-scale mapping capabilities of the MS-Mapping algorithm, we selected eight data sequences from the MS-Dataset, covering both indoor and outdoor environments. The total trajectory length is \SI{17.4}{km}, with a total collection time of \SI{126.6}{min}. Figure.~\ref{fig:8session} shows the incremental mapping results using these eight sessions, illustrating the final fused trajectory and the merged point cloud map. It is worth noting that point cloud noise caused by building glass, is not considered in this study. This further demonstrates its capabilities in large-scale environments and its flexibility in multi-session data fusion.

\subsection{Ablation Study}

The ablation study investigate the impact of various components on the performance of our MS system. We conducted experiments on the \texttt{RB2-PK1} sequences, using a voxel size of \SI{4.0}{m} and a map radius of \SI{200}{m}. In the WS module, frames with a distance exceeding \SI{1.8}{m} are considered keyframes, resulting in a keyframe ratio of 0.77.
The effectiveness of the proposed WS and UPGO components is evident from the quantitative results in Table~\ref{tab:ablation_experiment}. The map error visualization in Fig.~\ref{fig:ablation_map} further validates the importance of each module and their synergistic effect in our multi-session mapping system. 
The color bar in the figure ranges from red to green to blue, representing error levels from 0 to \SI{50}{cm}, with blue regions indicating lower errors. 
The presence or absence of the LC module has the most significant impact on map accuracy. When the UPGO and WS are introduced, map accuracy improves significantly, confirming their effectiveness in enhancing overall performance. These results, both quantitative and qualitative, highlight the importance of incorporating the WS, UPGO, and LC modules in the proposed system to achieve high-quality maps in multi-session scenarios.


\subsection{Run-time Analysis}\label{sub_sec:runtime}

We conducted two sets of experiments to analyze the system performance, ignoring the time cost for data preprocessing, such as reading the maps and graphs constructed from the old session.
The original pose graph for "\texttt{K1}" had 23986 edges, with 8.5\% being loop closure edges; "\texttt{B2}" had 29841 edges, with 7.1\% loop closure edges; and "\texttt{R1}" had 10243 edges, with 3.0\% loop closure edges. "\texttt{K1}" and "\texttt{B2}" represent incremental mapping experiments in larger scenarios (over \SI{11}{km}), while "R1" represents experiments in smaller scenarios (over \SI{2}{km}).

\begin{figure}
    \centering
    \includegraphics[width=0.45\textwidth]{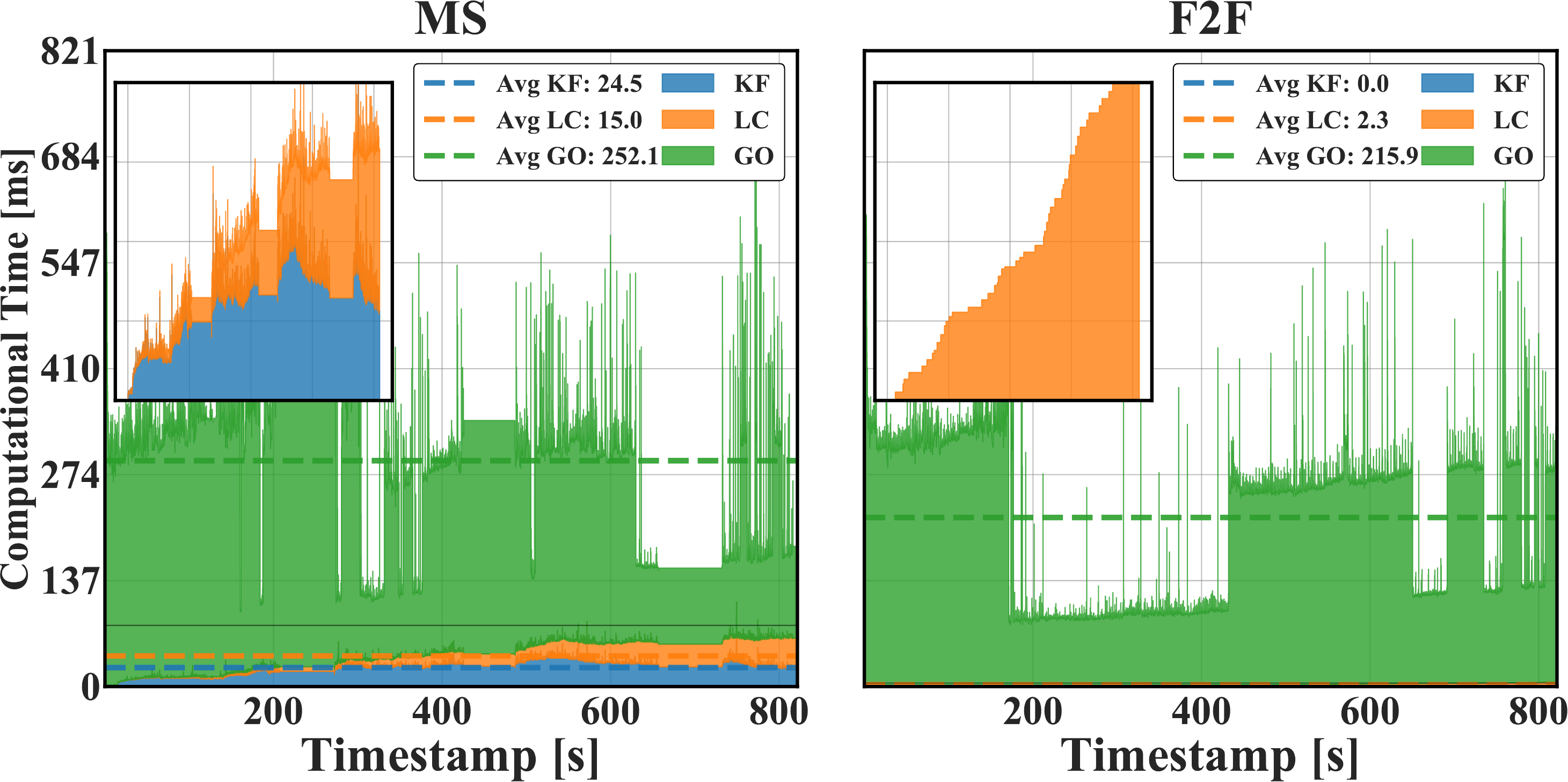}
    \caption{Performance Analysis and Comparison on \texttt{CP0}-\texttt{PK1}.}
    \label{fig:algorithm_time_analysis}
    \vspace{-1.5em}
\end{figure}

The first set of experiments focused on the runtime variation over time of key modules, including the keyframe selection module (KF), LC module, and Graph Optimization (GO) module. We compared these modules with F2F algorithm to analyze the system-level time consumption. 
Fig.~\ref{fig:algorithm_time_analysis} illustrates the results, with the left plot showing the doubled data for the LC and KF modules and the right plot increasing their data by a factor of 10 for better visibility. The KF module of the F2F algorithm consumes negligible time and is not shown in the right plot.
The results demonstrate that the KF module in MS-Mapping consumes less than 10\% of the total time, while the GO module has a higher consumption. However, the KF module reduces the overall graph size, resulting in the GO time consumption being nearly 18\% lower than the baseline, aligning with our expectations. The LC module, processed in a separate thread with a \SI{1}{Hz} frequency, is visualized using its average value at each timestamp.
For the second experiments, we quantitatively compared the average per-frame runtime of various modules, maintaining the same KFR. We set the voxel size of MS to \SI{5.0}{m}, the map radius to \SI{300}{m} (\textbf{MS1}) or \SI{200}{m} (\textbf{MS2}), and the Wasserstein distance threshold to \SI{2.5}{m}. For the baseline algorithm, we set the radius threshold to \SI{0.008}{m} (\textbf{F2F1}) and \SI{0.15}{m} (\textbf{F2F2}). 
Table~\ref{tab:algorithm_performance_comparison} presents the results.
The GO module is the most time-consuming in the whole system, despite using the incremental update algorithm ISAM2\cite{kaess2011}. Comparing MS1 and F2F1, under the same KFR, the single-frame time consumption of MS-Mapping increased by at least 30\%, with larger map sizes consuming more time. However, adjusting the map radius can mitigate this and improve performance.
The scenario of \texttt{R1} is much smaller than \texttt{K1} and \texttt{B2}. Comparing MS1 and MS2 reveals that slight changes in the KFR have minimal impact on performance in large scenes but are more significant in small scenes. This highlights the importance of considering scene scale when optimizing algorithm performance.
\begin{table}[t]
\centering
\caption{Average run-time [ms] comparison of key modules}
\renewcommand\arraystretch{1.0}
\label{tab:algorithm_performance_comparison}
\begin{threeparttable}
\begin{tabular}{l c c c c c c}
\toprule[0.03cm]
\textbf{Dataset} & \textbf{Method} & \textbf{KFR [\%]} & \textbf{KF} $\downarrow$ & \textbf{LC} $\downarrow$ & \textbf{GO} $\downarrow$ & \textbf{Total} $\downarrow$ \\
\midrule[0.03cm]
\multirow{4}{*}{\texttt{K1}} & 
\cellcolor{blue!50}MS1 & \cellcolor{blue!50}79.09 & \cellcolor{blue!50}24.47 & \cellcolor{blue!50}5.97 & \cellcolor{blue!50}240.75 & \cellcolor{blue!50}271.85 \\
& \cellcolor{blue!50}MS2 & \cellcolor{blue!50}98.65 & \cellcolor{blue!50}16.23 & \cellcolor{blue!50}5.25 & \cellcolor{blue!50}177.14 & \cellcolor{blue!50}200.20 \\
& F2F1 & 79.14 & 0 & 2.59 & 127.95 & 133.76 \\
& F2F2 & 84.47 & 0 & 1.33 & 166.27 & 172.38 \\
\midrule
\multirow{4}{*}{\texttt{B2}} & \cellcolor{blue!50}MS1 & \cellcolor{blue!50}70.70 & \cellcolor{blue!50}24.48 & \cellcolor{blue!50}14.99 & \cellcolor{blue!50}252.06 & \cellcolor{blue!50}284.83 \\
& \cellcolor{blue!50}MS2 & \cellcolor{blue!50}74.70 & \cellcolor{blue!50}18.22 & \cellcolor{blue!50}13.45 & \cellcolor{blue!50}234.97 & \cellcolor{blue!50}261.07 \\

& F2F1 & 70.90 & 0 & 2.62 & 219.47 & 226.34 \\
& F2F2 & 84.68 & 0 & 2.47 & 217.87 & 225.39 \\

\midrule
\multirow{2}{*}{\texttt{R1}} & \cellcolor{blue!50}MS1 & \cellcolor{blue!50}40.04 & \cellcolor{blue!50}32.47 & \cellcolor{blue!50}0.16 & \cellcolor{blue!50}36.75 & \cellcolor{blue!50}80.28 \\
& \cellcolor{blue!50}MS2 & \cellcolor{blue!50}80.35 & \cellcolor{blue!50}31.54 & \cellcolor{blue!50}1.17 & \cellcolor{blue!50}34.31 & \cellcolor{blue!50}72.89 \\


\bottomrule[0.03cm]
\end{tabular}
\footnotesize
\begin{tablenotes}[flushleft]
\item \texttt{K1}: \texttt{CP0}-\texttt{PK1} sequences; \texttt{B2}: \texttt{CP0}-\texttt{RB2} sequences; \texttt{R1}: \texttt{RB2}-\texttt{PK1} sequences.
\end{tablenotes}
\end{threeparttable}
      \vspace{-1.5em}
\end{table}

\subsection{Limitations}

The proposed MS-Mapping algorithm shows significant improvements in accuracy and robustness. However, there are several limitations that need to be addressed in future research.
\begin{enumerate}
    \item The incremental nature of MS-Mapping limits its ability to merge more than three session simultaneously. This necessitates additional data processing for large-scale map merging.
    \item The MS-Mapping system assumes known initial poses, restricting its operation to areas covered by existing maps. Integrating a place recognition algorithm is necessary to enhance its application in multi-robot SLAM.
    \item The computation of relative poses in uncertainty SLAM ignores pose correlations. There is insufficient evidence to support that considering pose correlations in large-scale LiDAR SLAM significantly improves accuracy.
    \item The approach relies solely on real-world experiments, where obtaining ground truth uncertainty is challenging. It lacks simulation experiments to quantitatively evaluate pose uncertainty in LiDAR SLAM.
    \item This MS-Mapping algorithm cannot handle loop closure outlier edges, which may cause to ill-pose optimization.
\end{enumerate}

%

\section{Conclusion}\label{sec:conclusion}

In this paper, we proposed MS-Mapping, a novel multi-session LiDAR mapping system that employs an incremental mapping scheme for accurate and robust map assembly in large-scale environments. Our approach introduced three key innovations: 1) a distribution-aware keyframe selection method to reduce data redundancy and pose graph complexity; 2) an adaptive uncertainty model for automatic least-squares adjustments without scene-specific parameter tuning; and 3) a comprehensive evaluation benchmark for the multi-session mapping task. 
Extensive experiments on public and self-collected data demonstrated the effectiveness of our new method. MS-Mapping showed superior performance in map accuracy and robustness compared to state-of-the-art methods. However, we acknowledge that MS-Mapping currently relies on a priori pose estimates from place recognition algorithms, which somewhat limits its broader application. 
Future work could focus on integrating place recognition capabilities directly into the system to eliminate this dependency. Additionally, enhancing the system performance to enable real-time crowdsourced mapping with multi-robot also presents an exciting avenue, potentially expanding the application scope of multi-session LiDAR mapping in dynamic, collaborative environments.

\bibliographystyle{IEEEtran}
\bibliography{ref}

\begin{IEEEbiography}[{\includegraphics[width=0.7in,height=0.9in,clip,keepaspectratio]{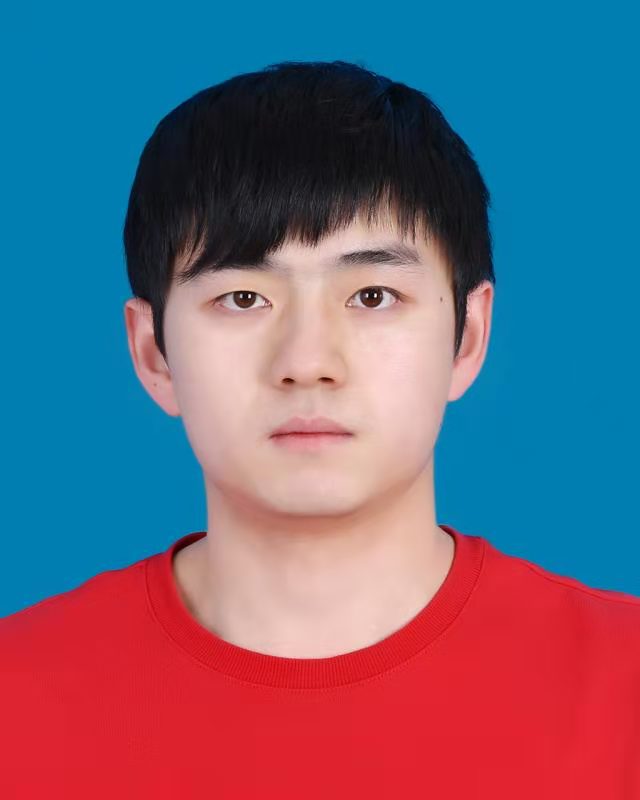}}]{Xiangcheng Hu} 
(Student Member, IEEE) received a B.Sc. degree from the North University of China, Taiyuan, China, in 2017 and the M.S. degree from Beihang University, Beijing, China, in 2020. He is currently working toward a Ph.D. degree with the Department of Electronic and Computer Engineering, Hong Kong University of Science and Technology, HKSAR.
\end{IEEEbiography}
\vspace{-1.0\baselineskip}

\begin{IEEEbiography}[{\includegraphics[width=0.7in,height=0.9in,clip,keepaspectratio]{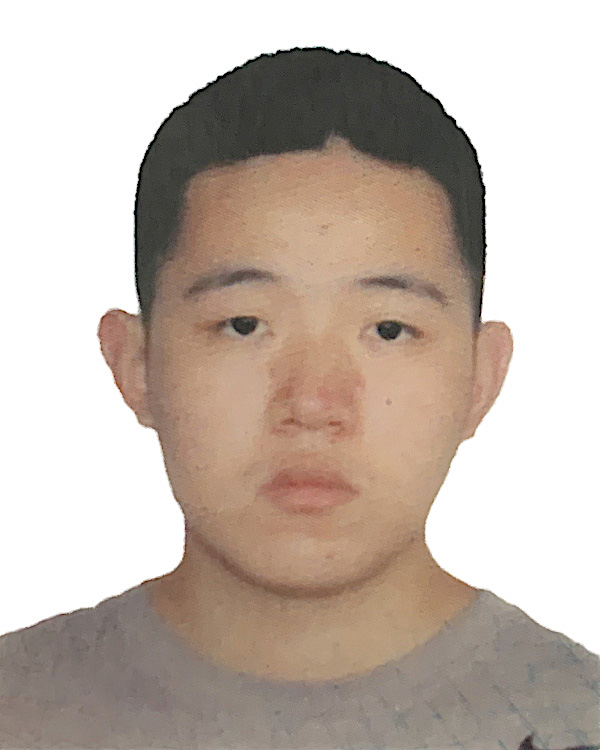}}]{Jin Wu} 
    received the B.S. degree from the University of Electronic Science and Technology of China, Chengdu, China. He is currently pursuing a Ph.D. degree in Electronic and Computer Engineering at the Hong Kong University of Science and Technology. From 2013 to 2014, he was an exchange student at KU Leuven. From 2019 to 2020, he worked at Tencent Robotics X, Shenzhen, China.
\end{IEEEbiography}

\begin{IEEEbiography}[{\includegraphics[width=0.7in,height=0.9in,clip,keepaspectratio]{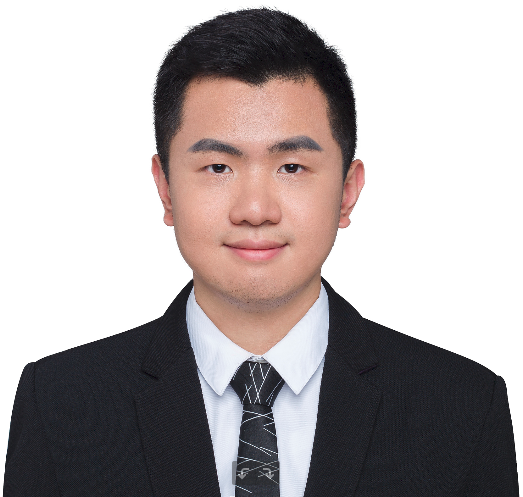}}]{Jianhao Jiao}
    (Member, IEEE) received his B.Eng. from Zhejiang University and his Ph.D. from HKUST in 2017 and 2021, respectively. He was a Research Associate at HKUST and is now a Senior Research Fellow at University College London. He is an Associate Editor for IEEE IROS 2024.
\end{IEEEbiography}
\vspace{-1.0\baselineskip}

\begin{IEEEbiography}[{\includegraphics[width=0.7in,height=0.9in,clip,keepaspectratio]{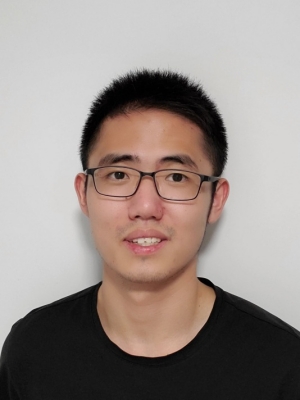}}]{Binqian Jiang}
    received the B.Eng degree in Information Engineering from Southeast University, Nanjing, China, in 2019. He is currently pursuing a Ph.D. in Electronic and Computer Engineering at the Hong Kong University of Science and Technology. His research interests include state estimation, sensor fusion, and SLAM.
\end{IEEEbiography}
\vspace{-1.0\baselineskip}

\begin{IEEEbiography}[{\includegraphics[width=0.7in,height=0.9in,clip,keepaspectratio]{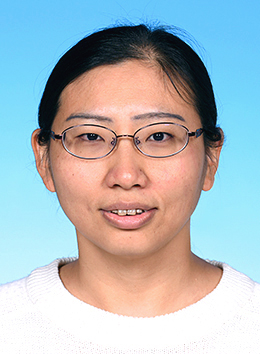}}]{Wei Zhang}
    (Member, IEEE) received her Ph.D. in Electrical Engineering from Princeton University, earning the Wu Prize. She joined HKUST in 2013 and is now a Full Professor. Her research includes reconfigurable systems, power management, embedded security, and emerging technologies. She has been general co-chair for ISVLSI 2018 and is an Associate Editor for IEEE and ACM journals, e.g., IEEE TCAD, IEEE TVLSI, ACM TECS.
\end{IEEEbiography}
\vspace{-1.0\baselineskip}



\begin{IEEEbiography}[{\includegraphics[width=0.7in,height=0.9in,clip,keepaspectratio]{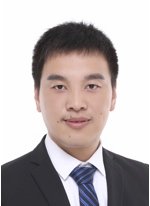}}]{Wenshuo Wang}
    (Member, IEEE) received his Ph.D. in mechanical engineering from Beijing Institute of Technology in 2018. He was a Research Assistant at UC Berkeley and the University of Michigan, and a Post-Doctoral Fellow at McGill, Carnegie Mellon, and UC Berkeley. He is now a Full Professor at BIT. He is on the editorial board of JFR, IEEE TIV, IEEE TVT.
\end{IEEEbiography}
\vspace{-1.0\baselineskip}

\begin{IEEEbiography}[{\includegraphics[width=0.7in,height=0.9in,clip,keepaspectratio]{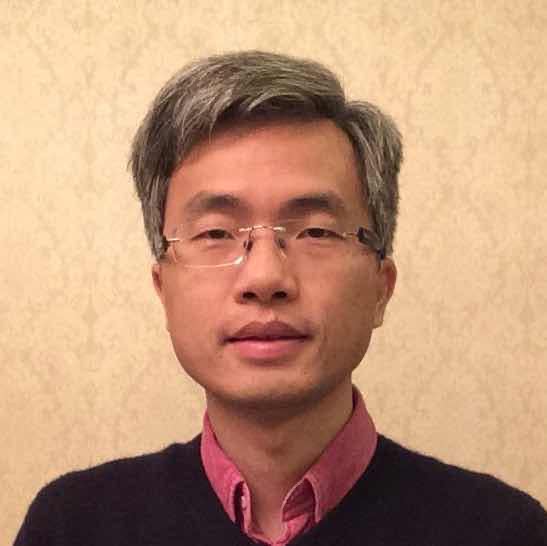}}]{Ping Tan}
  (Senior Member, IEEE) received his bachelor's degrees from Shanghai Jiao Tong University and his Ph.D. from HKUST. He is a Full Professor in Electronic and Computer Engineering at HKUST. His research includes computer vision, computer graphics, and robotics. He has served on editorial boards of IEEE TPAMI and as an area chair for conferences like CVPR, SIGGRAPH, and IROS.
\end{IEEEbiography}
\vspace{-1.0\baselineskip}

\end{document}